\documentclass[runningheads]{llncs}

\usepackage{eccv}

\usepackage{eccvabbrv}

\usepackage{graphicx}
\usepackage{booktabs}
\usepackage{color,soul}
\usepackage{colortbl}
\usepackage{todonotes}
\usepackage{tcolorbox}
\usepackage{graphicx}
\usepackage{booktabs}
\usepackage{multirow}
\usepackage{stackengine}
\usepackage{wrapfig}
\usepackage{enumitem}

\usepackage[accsupp]{axessibility}  

\usepackage{hyperref}

\usepackage{orcidlink}
\definecolor{mygreen}{HTML}{D1FFBD}

\begin{document}



\title{REVISION: Rendering Tools Enable Spatial Fidelity in Vision-Language Models} 

\titlerunning{REVISION}

\author{
Agneet Chatterjee \thanks{Equal contribution. Correspondence to \href{mailto:agneet@asu.edu}{agneet@asu.edu}.}\inst{1}\orcidlink{0000-0002-0961-9569} \and
Yiran Luo \textsuperscript{$\star$} \inst{1}\orcidlink{0000-0001-6533-8617} \and \\
Tejas Gokhale \inst{2}\orcidlink{0000-0002-5593-2804} \and 
Yezhou Yang \inst{1}\orcidlink{0000-0003-0126-8976} \and Chitta Baral\inst{1}\orcidlink{0000-0002-7549-723X}
}

\authorrunning{A. Chatterjee et al.}

\institute{Arizona State University \and University of Maryland, Baltimore County}

\maketitle

\begin{abstract}
  Text-to-Image (T2I) and multimodal large language models (MLLMs) have been adopted in solutions for several computer vision and multimodal learning tasks. 
  However, it has been found that such vision-language models lack the ability to correctly reason over spatial relationships. 
  To tackle this shortcoming, we develop the REVISION framework which improves spatial fidelity in vision-language models. 
  REVISION is a 3D rendering based pipeline that generates spatially accurate synthetic images, given a textual prompt. 
REVISION is an extendable framework, which currently supports 100+ 3D assets, 11 spatial relationships, all with diverse camera perspectives and backgrounds. Leveraging images from REVISION as additional guidance in a training-free manner consistently improves the spatial consistency of T2I models across all spatial relationships, achieving competitive performance on the VISOR and T2I-CompBench benchmarks. 
We also design RevQA, a question-answering benchmark to evaluate the spatial reasoning abilities of MLLMs, and find that state-of-the-art models are not robust to complex spatial reasoning under adversarial settings. Our results and findings indicate that utilizing rendering-based frameworks is an effective approach for developing spatially-aware generative models. 
Project Page : \url{https://agneetchatterjee.com/revision/}
\end{abstract}

\section{Introduction}
\label{intro}

Generative vision-language models \cite{dalle, geminiteam2023gemini} represent a significant step towards developing multimodal systems that bridge the gap between computer vision and natural language processing. 
Text-to-image (T2I) models \cite{stablediffusion,chen2023pixartalpha} convert text prompts to high-quality images, while multimodal large language models (MLLMs) \cite{llava,yang2023dawn} process images as inputs, and generate rich and coherent natural language outputs in response.  
As a result, these models have found diverse applications in robotics \cite{wake2023gpt4vision}, image editing \cite{hertz2022prompttoprompt}, image-to-image translation \cite{parmar2023zeroshot}, and more. 
However, recent studies \cite{kamath2023whats} and benchmarks such as DALL-Eval \cite{cho2023dall}, VISOR \cite{gokhale2023benchmarking}, and T2I-CompBench \cite{huang2023t2icompbench} have found that generative vision-language models suffer from a common mode of failure -- their inability to correctly reason over spatial relationships. 

We postulate that the lack of spatial understanding in generative vision-language models is a result of the lack of guidance from image-text datasets.
Compared to T2I models, graphics rendering tools such as Blender allow deterministic and accurate object placement, but are limited by their lower visual detail and photorealism and do not have intuitive workflows such as T2I models where users can generate images by simply typing a sentence.
To get the best of both worlds, in this work, we develop REVISION, a Blender-based image rendering pipeline which enables the synthesis of images with 101 3-dimensional object (assets), 11 spatial relationships, diverse backgrounds, camera perspectives, and lighting conditions. 
REVISION parses an input text prompt into assets and relationships and synthesizes the scene using Blender to exactly match the input prompt in terms of both objects and their spatial arrangement.

\begin{figure}[t]
    \centering 
    \includegraphics[width=\textwidth]{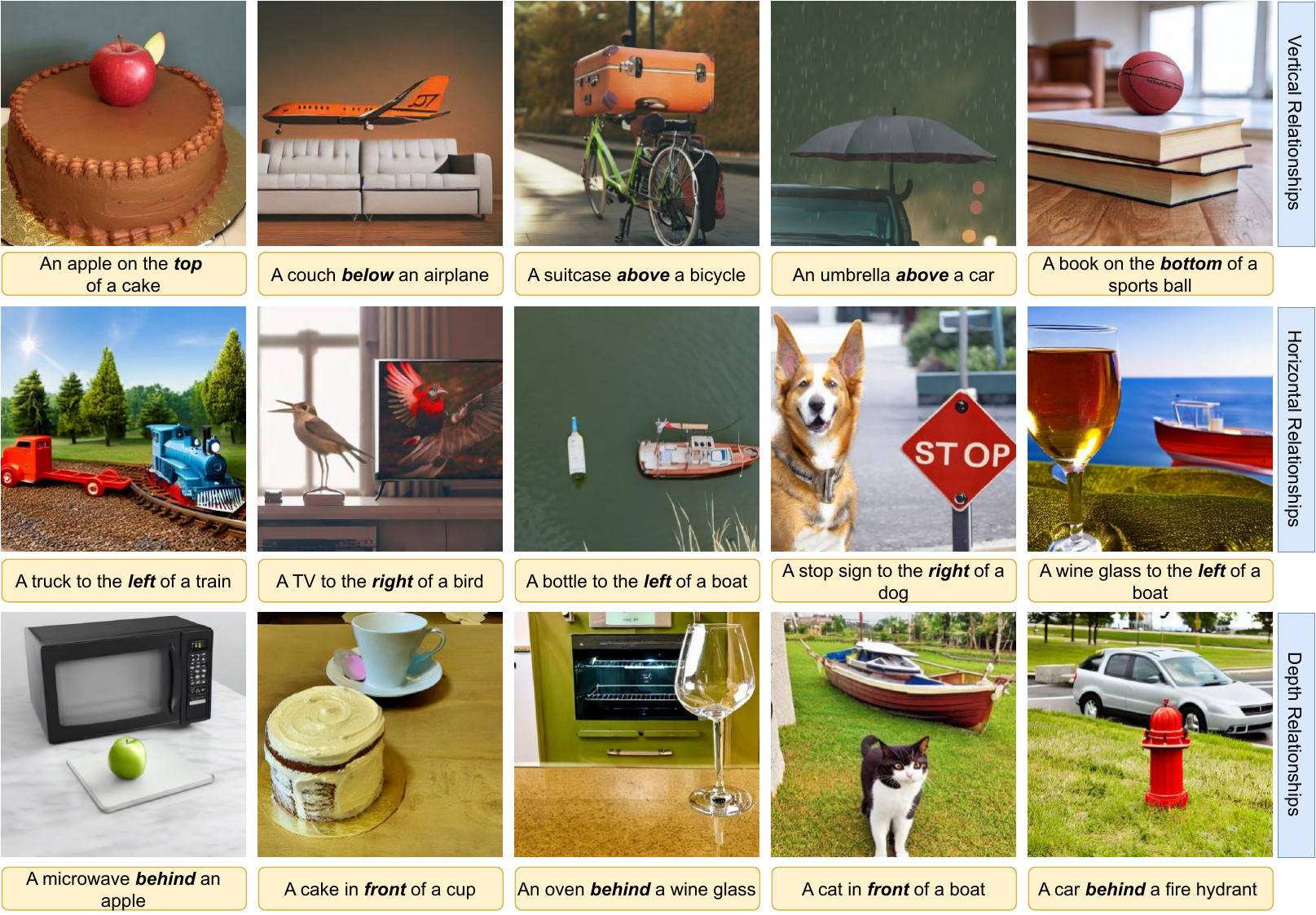}
    \caption{Text-to-Image models struggle to generate images that faithfully represent the spatial relationships mentioned in the input prompt. We develop REVISION, an efficient rendering pipeline that enables a training-free and guidance-based mechanism to address this shortcoming. 
    Our method results in improvements in spatial reasoning for T2I models for three dimensional relationships demonstrated by consistently higher scores on VISOR and T2I-CompBench benchmarks.
    }
    \label{fig:teaser_spade}
\end{figure}

In a \textit{training-free} manner, we leverage images from REVISION as additional guidance for existing T2I methods to their ability to generate spatially accurate images, and demonstrate improved performance on VISOR and T2I-CompBench benchmarks. 
We evaluate (i) the impact of utilizing diverse backgrounds from REVISION, (ii) the trade-off between controllability and photo-realism and (iii) the added generalization to complex prompts achieved by leveraging REVISION. For a holistic study, we introduce an extension to the VISOR benchmark, to include evaluation of depth relationships (in front of/behind). 

To assess the spatial and relational reasoning abilities of MLLMs, we also create the RevQA benchmark. We construct 16 diverse question types and their adversarial variations consisting of negations, conjunctions, and disjunctions. We perform holistic evaluations on 5 state-of-the-art MLLMs and 
discover significant shortcomings in their ability to accurately address complex spatial reasoning questions. These models also demonstrate a lack of robustness to adversarial perturbations, leading to a substantial decline in their performance.

The key contributions and findings are summarized below:
\begin{itemize}
    \item We develop the REVISION framework, a 3D rendering pipeline that is guaranteed to generate spatially accurate synthetic images, given an input text prompt. An extendable framework, REVISION currently accommodates 100+ assets across 11 spatial relationships and 3 diverse backgrounds, and support for multiple lighting conditions, camera perspectives, and shadows.
    \item We present an approach that utilizes images from REVISION in an efficient training-free manner, which results in improved spatial reasoning across multiple benchmarks. 
    Controlled experiments, ablations, and human studies reveal consistent improvements in generating images corresponding to the spatial relationships in the input prompt (as shown in Figure \ref{fig:teaser_spade}).
    \item We introduce the RevQA question-answering benchmark to evaluate spatial reasoning abilities of multimodal large language models. Our experiments reveal the shortcomings of state-of-the-art MLLMs in reasoning over complex spatial questions and their vulnerability to adversarial perturbations.
\end{itemize}

\section{Related Work}
\label{related_work}

\paragraph{\bf Generative Models for Image Synthesis.} 
Image generation and synthesis methods have advanced rapidly, progressing from early approaches such as generative adversarial networks (GAN) \cite{gan}, variational auto-encoders (VAE) \cite{vae}, and auto-regressive models (ARM) \cite{arm}, to contemporary text-to-image models including Stable Diffusion \cite{stablediffusion} and DALL-E \cite{ramesh2022hierarchical}. GLIDE \cite{nichol2022glide} adopts classifier-free guidance in T2I and explores the efficacy of CLIP \cite{clip} as a text encoder. Compared to GLIDE, Imagen \cite{imagen} adopts a frozen language model as the text encoder, reducing computational overhead, allowing for usage of large text-only corpus. 
Multiple variants of T2I models have been developed by leveraging T5-based text encoder \cite{chen2023pixartalpha}, T2I priors \cite{patel2023eclipse, razzhigaev2023kandinsky},  reward-based fine-tuning \cite{huang2023t2icompbench} and developing refiner models \cite{podell2023sdxl} for improved image-text alignment.

\paragraph{\bf Controllable Image Generation for Spatial Fidelity.}
To achieve better control over diffusion-based image synthesis, multiple methods have been proposed. ReCo \cite{reco}, GLIGEN \cite{gligen}, Control-GPT \cite{zhang2023controllable}, Composable Diffusion \cite{liu2023compositional} and ConPreDiff \cite{yang2024improving} all develop training-based methods to provide additional conditioning for T2I models. SPRIGHT \cite{chatterjee2024gettingrightimprovingspatial} introduces a spatially-focused large-scale dataset, by re-captioning 6 million images from existing vision datasets and demonstrate performance gains through an efficient training methodology. Test-time adaptations have also been proposed - {(i)} Layout Guidance \cite{chen2023trainingfree} restricts specific objects to their bounding box location through the modification of cross-attention maps; however it relies on bounding box annotations, {(ii)} LayoutGPT \cite{layoutgpt} and LLM-grounded Diffusion \cite{lian2024llm} leverage large language models (LLMs) to generate layouts and bounding box co-ordinates and, {(iii)} RealCompo \cite{zhang2024realcompo} combines multiple generative models for better spatial control. By developing an annotation-free cost-efficient framework we overcome the shortcomings of existing methods through REVISION. 

\paragraph{\bf Synthetic Images for Vision and Language.}
The flexibility and control provided during creation of synthetic images has led to various visuo-linguistic evaluation benchmarks using rendering tools.
CLEVR \cite{clevr} pioneered the utilization of synthetic objects in simulated scenes for visual compositionality reasoning. Many variants of CLEVR such as CLEVR-Hans \cite{stammer2021right}, CLEVR-Hyp \cite{sampat2021clevrhyp}, Super-CLEVR \cite{li2023super}, and CLEVRER \cite{clevrer} probe multiple facets of multimodal understanding with synthetic images and videos. 
PaintSkills introduced in DALL-EVAL \cite{cho2023dall} is an evaluation dataset that measures multiple aspects of a T2I model, which includes spatial reasoning, image-text alignment and social biases.

\paragraph{\bf Evaluation of Multimodal LLMs.} Multiple benchmarks have been proposed that evaluate reasoning capabilities of MLLMs. MMBench \cite{liu2023mmbench} evaluates models across 20 different dimensions, for a total of 2974 evaluation instances. The distinctive abilities of MLLMS to differentiate between coarse and fine-grained vision tasks is explored by MME \cite{fu2023mme} with images sourced from COCO. A limitation across all these benchmarks is that they collect instances from common VL datasets, increasing risk of data leakage \textit{and} do not evaluate spatial relationships at scale. RevQA fills this gap by developing a diverse set of synthetic and scalable image-question pairs for a holistic evaluation.
\section{The REVISION Framework} \label{spade_dataset}

\begin{figure}[t]
\centering
\includegraphics[width=\textwidth]{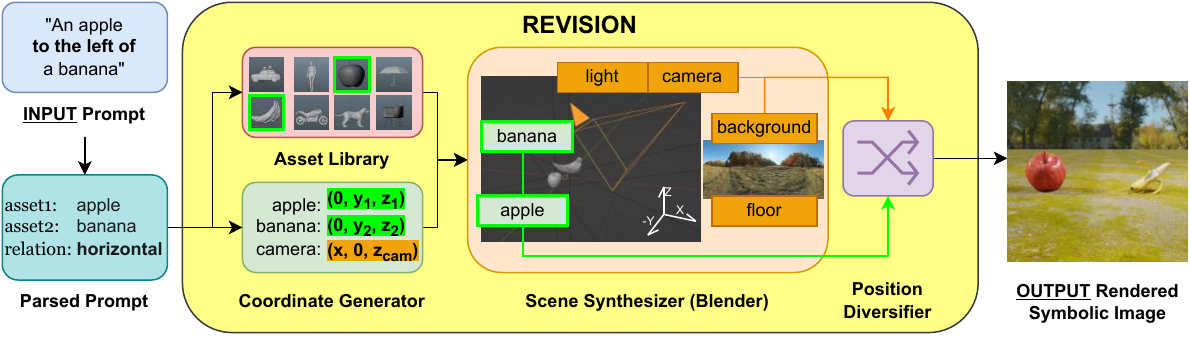}
\caption{REVISION parses a prompt into assets (objects) and the spatial relationship between them and synthesizes a symbolic image in Blender, placing the respective object assets at coordinates corresponding to the parsed spatial relationship.}
\label{fig:spade}
\end{figure} 

REVISION (Figure \ref{fig:spade}) is a rendering-based framework for generating spatially accurate images from an input prompt.
Given a prompt, we generate an image in Blender, where the two object 3D models and the camera view are situated according to the spatial relationship derived from the prompt. 
The components of REVISION are described below.

The \textbf{Asset Library} includes a large human-inspected collection of 3D models of realistic objects with variations in texture and shape.
Given an object name, the Asset Library randomly selects a matching asset rescaled to fit into a $1m$ cube to ensure that they are sufficiently visible in the final output. The Asset Library features 101 distinct classes of objects, 80 of which are from MS-COCO \cite{mscoco}. Each object class is associated with 3 to 5 royalty-free 3D model assets from \url{sketchfab.com}, with a total of 410 3D models. REVISION includes 3 background panoramas (Indoor, Outdoor, and White) from \url{polyhaven.com} and a corresponding textured floor asset from Sketchfab.

\begin{table}[t]
    \caption{Spatial relationships in REVISION and their rules for the Coordinate Generator. The objects are positioned from the camera's perspective.}
    \centering
    \footnotesize
    \begin{tabular}{lllr}
        \toprule
        Relation &Spatial Phrases &Coordinate Constraints&Distance (m) \\
        \midrule
        Horizontal &\textbf{{to the left, to the right }}&X and Z are 0. &[1, 1.5] \\
        Vertical &\textbf{{above, below, top, bottom }}&X and Y are 0. &[0.75, 1] \\
        Near &\textbf{{near, next to, on the side of}} &Z is 0. &[0.75, 1] \\
        Depth &\textbf{{in front of, behind }}&Z is 0. X$_{obj1}$=-X$_{obj2}$. &[1, 1.5] \\
        \bottomrule
    \end{tabular}
    \label{tab: revision-specs}
\end{table}

The \textbf{Coordinate Generator} deterministically generates 3D coordinates for the objects and the camera, given the names of the objects and the spatial relation extracted from the prompt.
As shown in Table \ref{tab: revision-specs}, REVISION supports four categories of spatial relationships between objects.
In our coordinate frame, the X-, Y-, and Z-axis represent depth, horizontal, and vertical relationships respectively.
To ensure that the objects are visible and the spatial relationship is obvious from the camera's view, the coordinate values for the objects on all three axes are confined within the range of $[-1m, 1m]$.
The camera is placed at $x=5m$ with its view always facing the origin point. The camera is at $z=2.5$ for depth relationships and at $z=1.5m$ otherwise.

The \textbf{Scene Synthesizer} assembles a 3D scene consisting of six main components: a camera, a light source, background, floor, and two objects. The two object assets and the camera are placed at their respective coordinates determined by the Coordinate Generator. Then the background asset, which is a 360-degree panorama image (modeled as a large sphere), is centered at the origin. The light source is added to a random position sufficiently higher than all objects in the scene. To prevent objects from appearing to float, the floor asset, a textured hyperplane orthogonal to the Z-axis, is positioned beneath the object asset with the lowest vertical coordinate. This floor placement also enables the object assets to cast shadows, enhancing the realism of the rendered image.

\begin{figure}[t]
    \centering
    \includegraphics[width=0.9\linewidth]{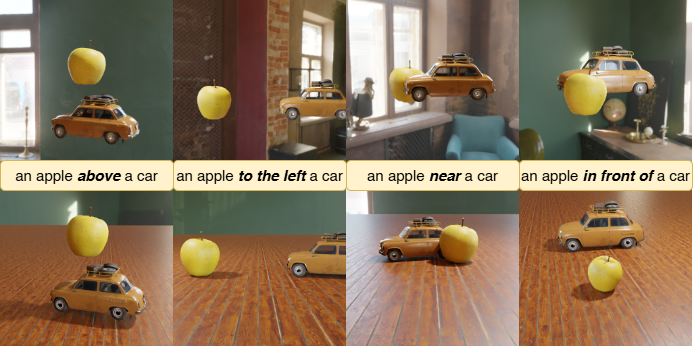}
    \caption{Outputs from the REVISION rendering pipeline for 4 spatial relationships types for identical assets, with (\textbf{bottom}) and without a floor (\textbf{top}).}
    \label{fig:spade_floor}
\end{figure} 

The \textbf{Position Diversifier} (Figure \ref{fig:spade_floor}) ensures diversity in object orientations, background, and the camera angles every time REVISION is invoked.
The background is rotated along the Z-axis, giving us a large number of static background options. In order to further diversify the perspective sizes and tilts of the object assets within the camera's view, we add random jitter to the position and orientation of the camera. We also add random small rotations to the objects along the Z-axis and vary the distance between the objects so that they are not always symmetric around the origin.
See Supplementary Materials for more details.

\section{Improving Spatial Fidelity in T2I Generation}\label{T2I}

\subsection{Training-Free Image Generation with REVISION} \label{method}

\begin{figure}[t]
    \centering
    \includegraphics[width=\linewidth]{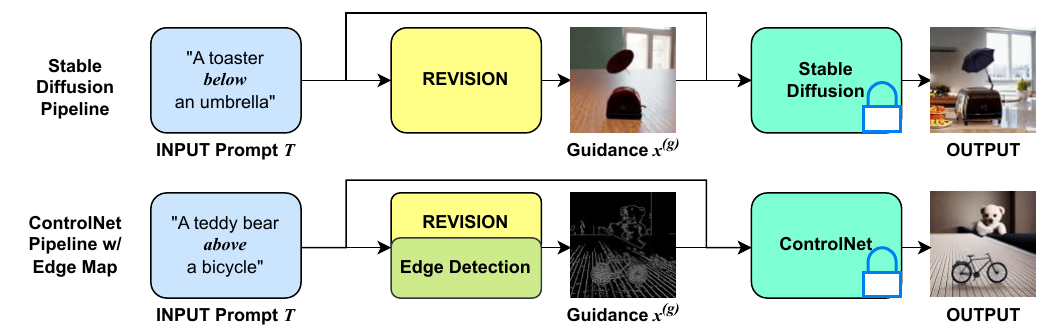}
    \caption{Given a user-provided input prompt $T$, we generate a corresponding synthetic image $x^{(g)}$ using REVISION. With input prompt $T$ and guidance $x^{(g)}$, we perform training-free image synthesis based on existing T2I pipelines such as Stable Diffusion or ControlNet to obtain a spatially accurate image.}
\label{fig:method}
\end{figure}

Given an input prompt ($T$) , we first generate a spatially accurate reference image  ($x^{(g)}$) leveraging our REVISION pipeline. We then perform training-free image synthesis to generate an image $I$, i.e. $\phi(I | x^{(g)}, T )$, where $\phi$  is a T2I model. We re-formulate the standard text-to-image pipeline into an image-to-image pipeline, conditioned by text, as shown in Figure \ref{fig:method}.

Standard diffusion methods such as Stable Diffusion (SD) generate an image by iteratively de-noising a Gaussian noise vector. Stochastic Differential Editing (SDEdit) \cite{meng2022sdedit}, on the other hand, starts from a guide image ($x^{(g)}$, in our case), adds Gaussian noise to it, and denoises it to produce the synthesized image $I$. We use SDEdit within our Stable Diffusion pipeline and perform image generation guided by $x^{(g)}$. We also explore the ControlNet \cite{controlnet} backbone, which allows fine-grained control over SD. Using ControlNet allows us to address two key points: \textbf{a)} our reference images provide enough spatial information even when low-level features are extracted from them and, \textbf{b)} we can mitigate any attribute-related biases present in the assets. 

\subsection{Experimental Setup}

We study the efficacy of REVISION on two widely accepted benchmarks for spatial relationship, VISOR \cite{gokhale2023benchmarking} and T2I-CompBench \cite{huang2023t2icompbench}, which have 25,280 and 300 spatial prompts, respectively. For each evaluation prompt in the respective benchmarks, we generate a corresponding image from our REVISION pipeline and perform training-free image generation as described in Section \ref{method}.

We leverage 3 variants of Stable Diffusion (SD), versions 1.4,
1.5, and 2.1 as our baseline models.
For ControlNet, we use the canny edge-conditioned SD model. For holistic evaluations, we also report the Inception Score (IS) \cite{salimans2016improved} where applicable. For all subsequent tables, the \textbf{bold} values denote the best performance while \underline{underlined} values indicate the second-best performance. 

\subsection{Results and Analysis}

\begin{figure}[!t]
    \centering 
    \includegraphics[width=.9\textwidth]{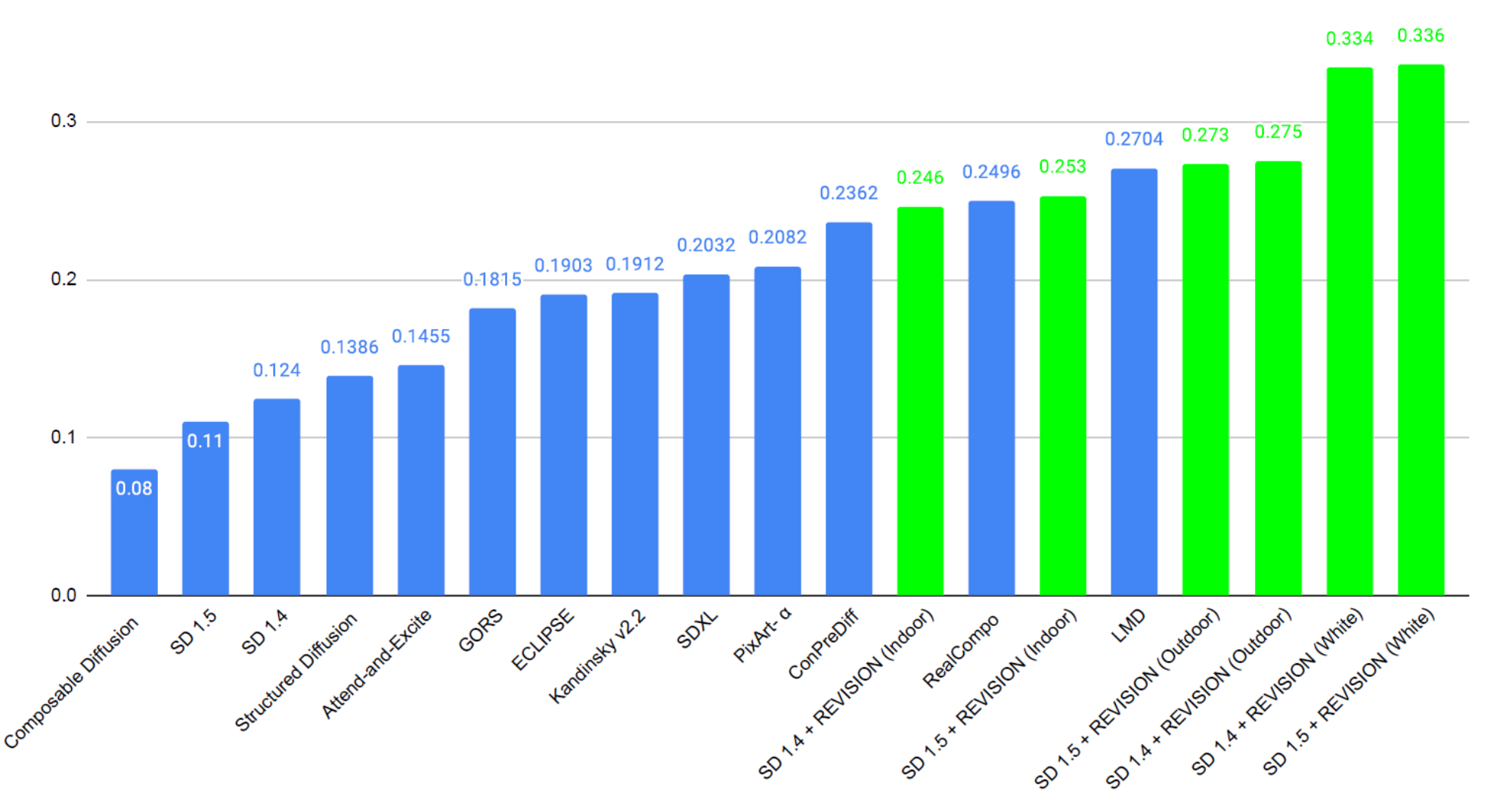}
    \caption{Comparing the T2I-CompBench spatial scores of REVISION-based guidance (green) with other leading T2I models and methods (blue).}
    \label{fig:t2i_compbench_other_methods}
\end{figure}
\begin{table}[t]
    \centering
    \footnotesize
    \caption{The incorporation of REVISION as a guiding framework significantly enhances the spatial reasoning performance of Stable Diffusion (SD) models. Results highlighted in \textcolor{ForestGreen}{green} represent scores achieved with images from REVISION.}
    \begin{tabular}{@{}l rr r rrrr@{}}
        \toprule
        \multirow{2}{*}{\textbf{Method}} & \multirow{2}{*}{\textbf{OA (\%)}} & \multicolumn{6}{c}{\textbf{VISOR (\%)}} \\
         \cmidrule{3-8}
        & & \textbf{uncond} & \textbf{{cond}} & \multicolumn{1}{c}{\textbf{1}} &  \multicolumn{1}{c}{\textbf{2}} &  \multicolumn{1}{c}{\textbf{3}} &  \multicolumn{1}{c}{\textbf{4}}\\
        \midrule
        SD 1.4         & 29.86 & 18.81 & 62.98 & 46.60 & 20.11 &  6.89 & 1.63 \\
        \rowcolor{mygreen}\quad + REVISION & 53.96 & 52.71 & 97.69 & 77.79 & 61.02 & 44.90 & 27.15 \\ 
        \midrule
        SD 1.5         & 28.43 & 17.51 & 61.59 & 44.27 & 18.12 & 6.28 & 1.35 \\
        \rowcolor{mygreen}\quad + REVISION & 54.33 & {53.08} & 97.72 & {78.07} & {61.27} & {45.44} & {27.55} \\ 
        \midrule
        SD 2.1         & 47.83 & 30.25 & 63.24 & 64.42 & 35.74 & 16.13 &  4.70  \\
        \rowcolor{mygreen}\quad + REVISION & 48.26 & 47.11  & 97.61 & 76.07 & 55.75 & 37.10 & 19.53 \\
        \bottomrule
    \end{tabular}
    \label{tab:visor_baseline_comp}
\end{table}
\begin{table}[!t]
    \centering
    \footnotesize
    \caption{\textbf{Results on the VISOR Benchmark}. With REVISION, we consistently outperform existing T2I methods on the VISOR benchmark.}
    \begin{tabular}{@{}l rr r rrrr@{}}
        \toprule
        \multirow{2}{*}{\textbf{Method}} & \multirow{2}{*}{\textbf{OA (\%)}} & \multicolumn{6}{c}{\textbf{VISOR (\%)}} \\
         \cmidrule{3-8}
        & & \textbf{uncond} & \textbf{{cond}} & \multicolumn{1}{c}{\textbf{1}} &  \multicolumn{1}{c}{\textbf{2}} &  \multicolumn{1}{c}{\textbf{3}} &  \multicolumn{1}{c}{\textbf{4}}\\
        \midrule
        GLIDE   \cite{nichol2022glide}       &  3.36 &  1.98 & 59.06 &  6.72 & 1.02 & 0.17 & 0.03 \\ 
        DALLE-mini \cite{Dayma_DALL·E_Mini_2021}    & 27.10 & 16.17 & 59.67 & 38.31 & 17.50 &  6.89 & 1.96 \\
        DALLE-v2  \cite{ramesh2022hierarchical}     & \textbf{63.93} & 37.89 & 59.27 & 73.59 & 47.23 & 23.26 & 7.49 \\
        Layout Guidance \cite{chen2023trainingfree} & 40.01 & 38.80 & 95.95 & - & - & - & - \\
        Control-GPT \cite{zhang2023controllable} & 48.33 & 44.17 & 65.97 & 69.80 & 51.20 & 35.67 & 20.48 \\
        Structured Diffusion \cite{feng2023trainingfree} & 28.65 & 17.87 & 62.36 & 44.70 & 18.73 & 6.57 & 1.46 \\ 
        Attend-and-Excite \cite{chefer2023attendandexcite} & 42.07 & 25.75 & 61.21 & 49.29 & 19.33 & 4.56 & 0.08 \\ 
        \midrule
        SD 1.4 + REVISION & 53.96 & 52.71 & \underline{97.69} & 77.79 & 61.02 & 44.90 & 27.15 \\
        SD 1.5 + REVISION & 54.33 & \underline{53.08} & \textbf{97.72} & \underline{78.07} & \underline{61.27} & \underline{45.44} & \underline{27.55} \\
        SD 2.1 + REVISION & 48.26 & 47.11  & 97.61 & 76.07 & 55.75 & 37.10 & 19.53 \\
        ControlNet + REVISION & \underline{56.88} & \textbf{55.48}  & 97.54  & \textbf{78.82} & \textbf{62.93} & \textbf{48.58} & \textbf{31.59} \\
        \bottomrule
    \end{tabular}
    \label{tab:visor_main}
\end{table}
\begin{figure}[t]
    \centering
    \begin{minipage}{0.48\textwidth}
        \centering
        \includegraphics[width=1\textwidth]{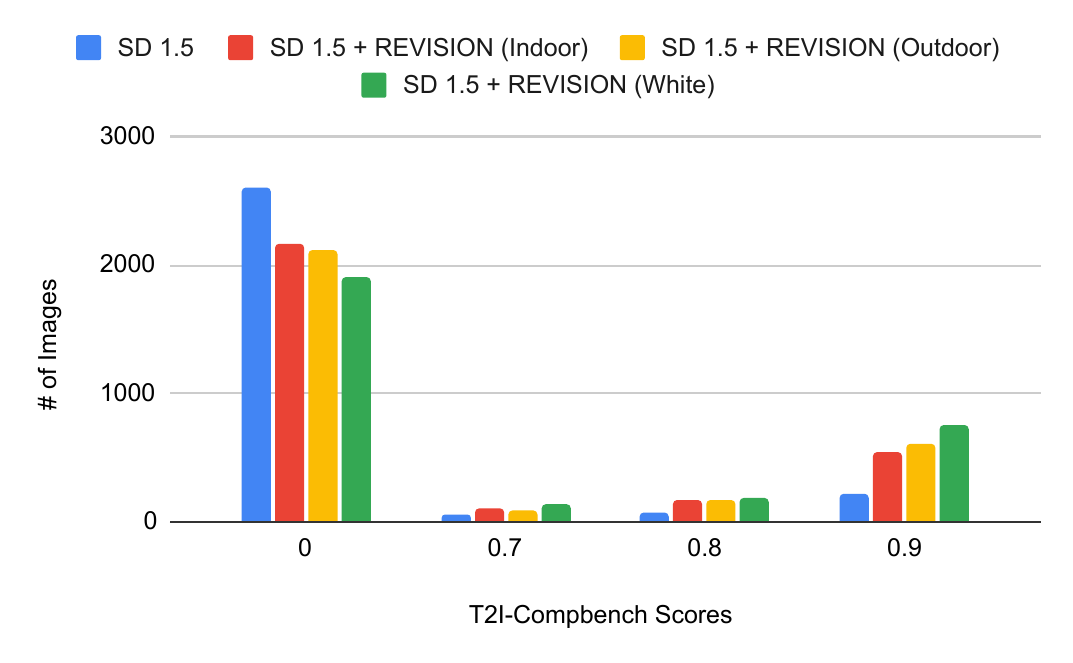}
        \caption{REVISION improves T2I-CompBench Spatial Score (0 indicates missing objects, 1 denotes perfect object generation and spatial accuracy.)}%
     \label{fig:t2i_compbench}
    \end{minipage}\hfill
    ~
    \begin{minipage}{0.48\textwidth}
        \centering
        \includegraphics[width=1\textwidth]{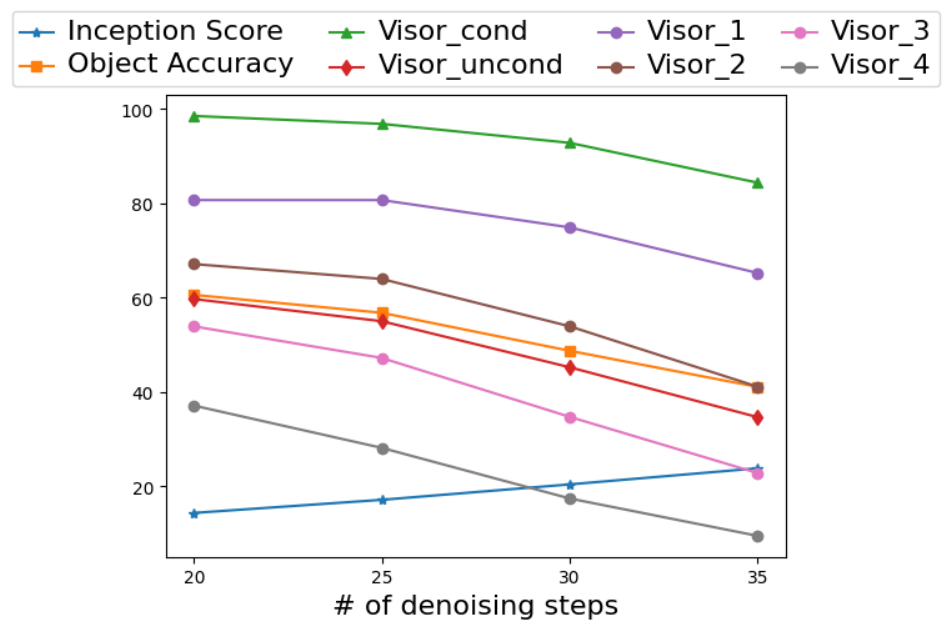}
        \caption{Benchmarking the trade-off between spatial accuracy (VISOR) and Inception Score, achieved with REVISION.}%
     \label{fig:chart_denoising}
    \end{minipage}
\end{figure}
\textbf{Improvements over Baseline Models -} We summarize our representative improvements over the baseline and existing methods, on the VISOR and T2I-CompBench benchmarks in Table \ref{tab:visor_baseline_comp} and Figure \ref{fig:t2i_compbench_other_methods} respectively. The results in Table  \ref{tab:visor_baseline_comp} are shown with reference images on a white background and \# of denoising steps = 30. As shown in Table \ref{tab:visor_baseline_comp}, we improve on all aspects of spatial relationships compared to our baseline methods. On SD 1.5, we achieve a \textbf{91.1\%} improvement in Object Accuracy (OA) and a \textbf{58.6\%} improvement on the conditional score. Specifically, we generate objects more accurately and achieve a high \% of accuracy when spatially synthesizing them in the image. Interestingly, through REVISION, we increase the likelihood of consistently generating spatially correct images, as can be seen by the relatively high value of VISOR$_4$. On VISOR (Table \ref{tab:visor_main}), REVISION enables baseline Stable Diffusion models to consistently outperform existing methods, across all aspects. Compared to the best open-source model, Control-GPT, we achieve a $\Delta$ improvement of \textbf{17.69\%},  \textbf{48.12\%}, and \textbf{25.6\%} on OA,  VISOR$_{cond}$, and  VISOR$_{uncond}$ respectively.

On T2I-CompBench (Figure \ref{fig:t2i_compbench_other_methods}), we observe similar improvement trends across diverse backgrounds, with baseline models guided by REVISION achieving consistent performance gains on the benchmark. In addition to enhancing spatial accuracy, REVISION improves prompt fidelity by ensuring that images contain all objects mentioned in the input prompt (Figure \ref{fig:t2i_compbench}).

\noindent\textbf{Consistent Performance Across Relationship Types - } Across all spatial relationship types, REVISION achieve  a consistently high performance score across the VISOR metrics as shown in Table \ref{tab:split_by_rel}; a shortcoming prevalent in other methods. For example, the largest deviation in VISOR$_{cond}$ performance for ControlNet + REVISION is 0.21\% between \textit{left} and \textit{below} relationships; in comparison Control-GPT deviates as much as 6.8\% for the same.

\begin{table}[t]
    \centering
    \footnotesize
    \caption{VISOR\textsubscript{cond} and Object Accuracy, split across relationship types. $\sigma_\texttt{Vc}$ and $\sigma_\texttt{OA}$ denote the respective metric's standard deviation w.r.t the relationships. Regardless of the spatial relation, REVISION enables T2I models to consistently produce spatially accurate images, a challenge faced by earlier approaches.}
    \begin{tabular}{@{}l rrrrr c rrrrr@{}}
        \toprule
        \multirow{2}{*}{\textbf{Method}} & \multicolumn{5}{c}{\textbf{VISOR\textsubscript{cond} (\%)}} & \hphantom & \multicolumn{5}{c}{\textbf{Object Accuracy (\%)}}\\
        \cmidrule{2-6}\cmidrule{8-12}
                    & left & right & above & below  & $\sigma_\texttt{Vc}$                       && left & right & above & below                            & $\sigma_\texttt{OA}$  \\\midrule
                    GLIDE       & 57.78 & 61.71 & 60.32 & 56.24 & 2.46             && 3.10 &     3.46 &      3.49 &            3.39                                     & 0.18 \\
                    DALLE-mini  & 57.89 & 60.16 & 63.75 & 56.14 & 3.29             && 22.29 &    21.74 &     33.62 &          30.74                                     & 5.99 \\
                    DALLE-v2    & 56.47 & 56.51 & 60.99 & 63.24 & 3.38             && \textbf{64.30} &    \textbf{64.32} &     \textbf{65.66} &          \textbf{61.45} & 1.77 \\
                    Control-GPT & 72.50 & 70.28 & 67.85 & 65.70 & 2.95             && 49.80 & 48.27 & 47.97 & 46.95                                                     & 1.18 \\
                    \midrule
                    SD 1.4 + REVISION & 97.53 & 97.45 & \textbf{98.09} & 97.66                                    & 0.29  && 52.42 & 52.11 & 56.93 & 54.38                                                      & 2.22\\
                    SD 1.5 + REVISION & \underline{97.57} & \textbf{97.53} & \underline{98.05} & \underline{97.70}& 0.24  && 52.99 & 52.59 & 56.80 & 54.92                                                      & 1.94\\
                    SD 2.1 + REVISION & \textbf{97.81} & \underline{97.46} & 97.91 & 97.28                        & 0.30  && 46.70 & 47.94 & 49.70 & 48.71                                                      & 1.27\\
                    ControlNet + REVISION & 97.51 & 97.25 & 97.65 & \textbf{97.72}                                & 0.21  && \underline{55.10} & \underline{55.14} & \underline{58.98} & \underline{58.29}      & 2.05\\
        \bottomrule
    \end{tabular}
    \label{tab:split_by_rel}
\end{table}

\subsection{Ablation Studies}

\begin{figure}[t]
    \centering 
    \includegraphics[width=\linewidth]{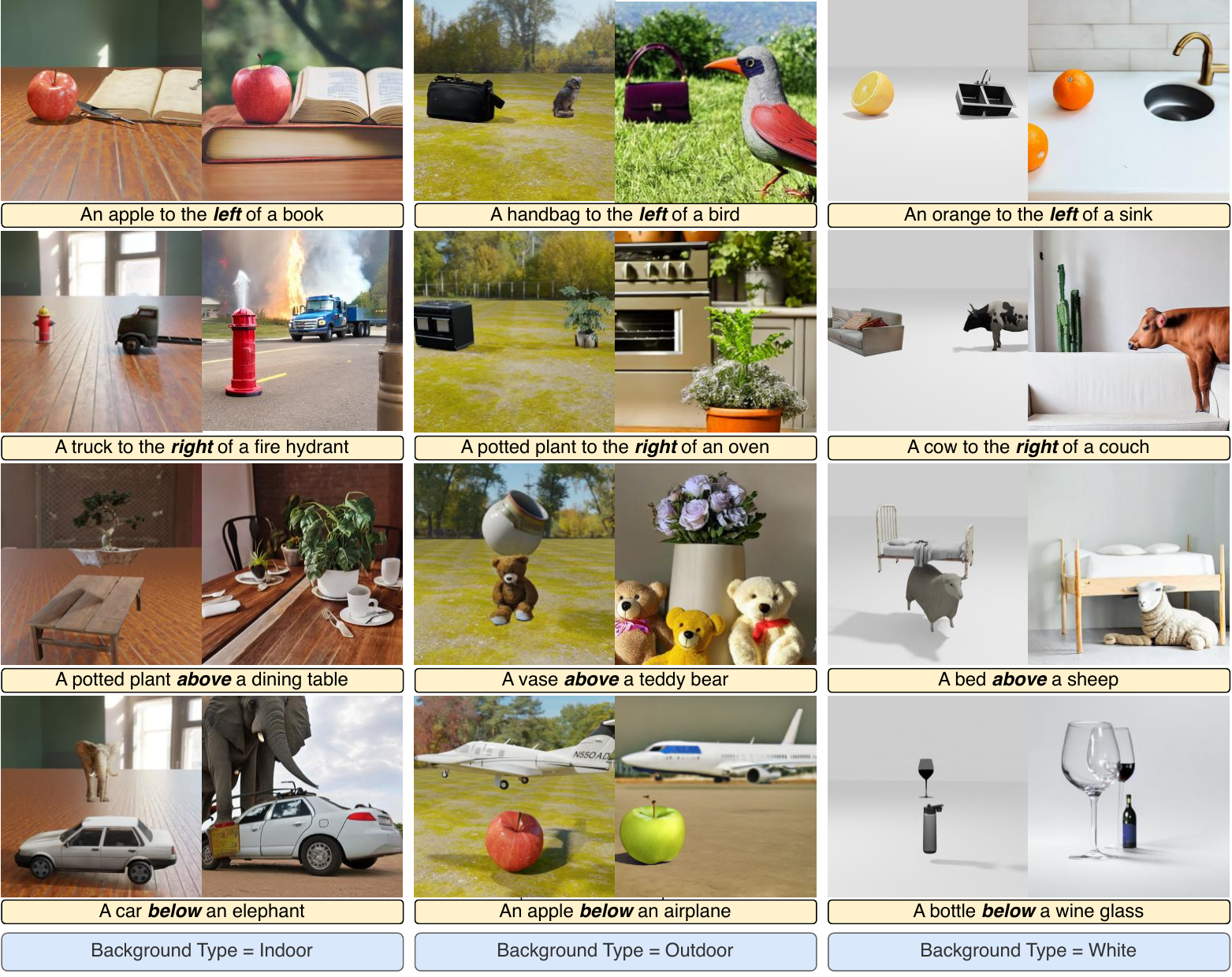}
    \caption{Illustrative examples depicting the variation of generated images across the three variants of backgrounds in REVISION. For each pair, the image on the left is from REVISON and the image on the right is generated from the T2I model.}%
    \label{fig:bg_impact}
\end{figure}

\begin{table}[t]
    \caption{The impact of the 3 background types in the REVISION pipeline on the VISOR benchmark. While best performance is achieved with a white background, diverse outputs are attained with the outdoor background type.}
    \centering
    \footnotesize
    \begin{tabular}{@{}cc rr rrrrrr@{}}
        \toprule 
        \multirow{2}{*}{\textbf{Model}}  & \multirow{2}{*}{\textbf{Background}} & \multirow{2}{*}{\textbf{IS} ($\uparrow$)} & \multirow{2}{*}{\textbf{OA (\%)}} & \multicolumn{6}{c}{\textbf{VISOR (\%)}} \\
         \cmidrule{5-10}
         & & & & \textbf{uncond} & \textbf{{cond}} & \textbf{1} & \textbf{2} & \textbf{3} & \textbf{4}\\
        \midrule
        &White          &16.16&  \underline{53.96} & \underline{52.71} & \underline{97.69} & \underline{77.79} & \underline{61.02} & \underline{44.9} & \underline{27.15} \\
        SD 1.4&Indoor     &19.11& 48.53 & 45.12 & 92.97 & 74.82 & 53.79 & 34.78 & 17.09 \\
        &Outdoor       &\textbf{20.16}& 44.32 & 41.80 & 94.31 & 69.79 & 49.38 & 31.86 & 16.17 \\
        \midrule
        &White          &16.27&  \textbf{54.33} & \textbf{53.08} &  \textbf{97.72} & \textbf{78.07} & \textbf{61.27 }& \textbf{45.44} & \textbf{27.55} \\
        SD 1.5&Indoor     &19.11& 48.77 & 45.28 & 92.85 & 74.93 & 53.96 & 34.77 & 17.47 \\
        &Outdoor       & \underline{19.66} & 43.99 & 41.51 & 94.36 & 69.48 & 48.58 & 31.46 & 16.52 \\
        \midrule
        &White          &12.79&  48.26 & 47.11 & 97.61 & 76.07 & 55.75 & 37.10 & 19.53 \\
        SD 2.1&Indoor     &11.52& 31.08 & 29.37 & 94.50 & 59.80 & 33.96 & 17.40 & 6.34 \\
        &Outdoor       &10.51& 36.37 & 34.67 & 95.34 & 65.05 & 41.23 & 23.05 & 9.36 \\
        \bottomrule
    \end{tabular}
    \label{tab:bg_table}
\end{table}
\textbf{Impact of Background - } In Table \ref{tab:bg_table}, we enumerate the impact of the background types in the images from the REVISION pipeline and the downstream trade-off between VISOR performance and model diversity. Utilizing white backgrounds that exclusively feature the two objects in question minimizes potential distractions for the model. Conversely, when the model is presented with initial reference images incorporating indoor or outdoor backgrounds, it exhibits the capacity to identify and leverage \textit{distractor} objects, resulting in the generation of diverse images. As shown in Figure \ref{fig:bg_impact}, all generated images maintain spatial accuracy, but noisier reference images result in greater diversity.

\begin{figure}[t]
    \centering 
    \includegraphics[width=\linewidth]{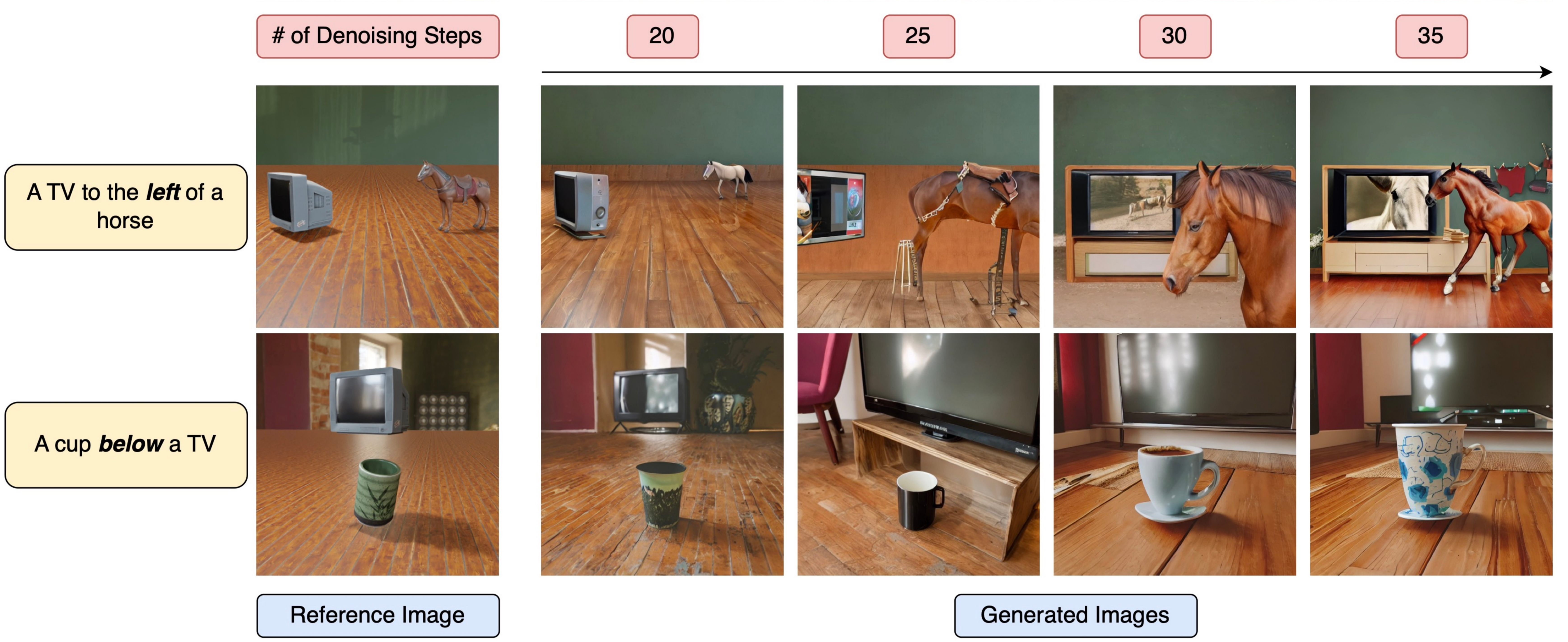}
    \caption {Illustrative examples showing the trade-off between photo-realism and denoising steps, while maintaining generating spatially accurate images using REVISION.}%
    \label{fig:images_denoising}
\end{figure}

\paragraph{\bf Controllability vs Photo-Realism - } In this setup, we study the impact of the \# of denoising steps and its trade-off with photo-realism. As shown in Figure \ref{fig:chart_denoising} that while the performance on VISOR deteriorates with additional \# of denoising steps, it improves the model's ability to be more diverse and photo-realistic. In Figure \ref{fig:images_denoising}, we demonstrate that by utilizing REVISION, baseline models can preserve their spatial coherency while iteratively demonstrating a higher degree of photo-realism, through more \# of denoising steps.

\subsection{Extending VISOR for Depth Relationships}
We further extend the VISOR benchmark for Depth relationships (in front of/behind). We utilize Depth Anything\cite{depthanything} for generation of depth maps and OWLv2\cite{owlv2} for object detection. Given a T2I generated image $I$ and its prompt $T$ that contain two objects $o_1, o_2$, we obtain its depth map $I_D$ using Depth Anything. We then retrieve the centroids detected for the two objects $c_{o_1}, c_{o_2}$  using OWLv2. At these centroid coordinates, we acquire the depth values for the two objects from the depth map $I_D(c_{o_1}), I_D(c_{o_2})$. We check if the acquired depth values match the spatial relationship in the prompt, and evaluate similar to VISOR. As shown in Table \ref{tab:visor_depth_comp}, REVISION improves VISOR scores across all metrics and across multiple denoising steps. 
\begin{table}[t]
    \caption{Comparing baseline methods against REVISION-guided image synthesis on depth relationships. DS denotes the \# of Denoising Steps. }
    \centering
    \begin{tabular}{@{}l rr rrrr@{}}
        \toprule
        \multirow{2}{*}{\textbf{Setting}} & \multirow{2}{*}{\textbf{OA (\%)}} & \multicolumn{5}{c}{\textbf{VISOR (\%)}} \\
         \cmidrule{3-7}
        & & \textbf{uncond}  & \multicolumn{1}{c}{\textbf{1}} &  \multicolumn{1}{c}{\textbf{2}} &  \multicolumn{1}{c}{\textbf{3}} &  \multicolumn{1}{c}{\textbf{4}}\\
        \midrule
        SD 1.5 (baseline) & 41.52       & 27.15 &  60.12   &   31.82    &  13.19     &  3.47 \\
        \quad + REVISION (DS=30)   &  \textbf{58.32}    & 29.62 & 64.74   &   34.53  &    14.99   &    4.22   \\
        \quad + REVISION (DS=35)   &   \underline{52.05}  & \textbf{32.08} & \textbf{68.11 }   &  \textbf{37.92}    &  \textbf{17.51}   &    \textbf{4.78} \\
        \quad + REVISION (DS=40)   &  47.43   & \underline{30.46}  & \underline{65.50}  &    \underline{35.99}    &  \underline{15.74}     &  \underline{4.64}  \\
        \bottomrule
    \end{tabular}
    \label{tab:visor_depth_comp}
\end{table}

\subsection{Human Evaluations} \label{human_studies}
To verify the generalizability of REVISION-based guidance on T2I models, we perform 2 distinct experiments and conduct human evaluations for validation. For each experiment, we independently sample 200 generated images and take the average scores across 4 workers. We also report unanimous (100\%) and majority (75\%) agreements between the workers for each experiment. 

\paragraph{\bf Prompts of Multiple Objects and Relationships -} In this experiment, we generate reference images using prompts that include 2 spatial phrases and 3 objects, and use these images to guide T2I generation. Each generated image is evaluated for accuracy based on the input spatial prompts. We achieve an accuracy of 79.62\% when at least 1 phrase is correctly represented in the image and 46.5\% when both phrases are correctly represented. The unanimous and majority agreements among evaluators are 64.5\% and 86.5\%, respectively.

\paragraph{\bf Out-of-Distribution Objects -} We consider prompts containing exactly one object not found in the REVISION Asset Library. Given a prompt that mentions an OOD object, we find the semantically closest object (list in Supplementary  Material) in our library and use their corresponding image as guidance. For example, we generate an image of ``a \textbf{\textit{helicopter}} above a bicycle'' by providing a reference image of ``an \textbf{\textit{airplane}} above a bicycle''. An accuracy of  63.62\% is found with an unanimous and majority agreement of 67\% and 90.5\%, respectively.
\section{RevQA: A Spatial Reasoning Benchmark for MLLMs}
We leverage the determinism of the REVISION pipeline to construct a new visual question answering benchmark (RevQA) for evaluating the spatial reasoning abilities of multimodal large language models.

\begin{figure}[t]
    \centering 
    \includegraphics[width=\linewidth]{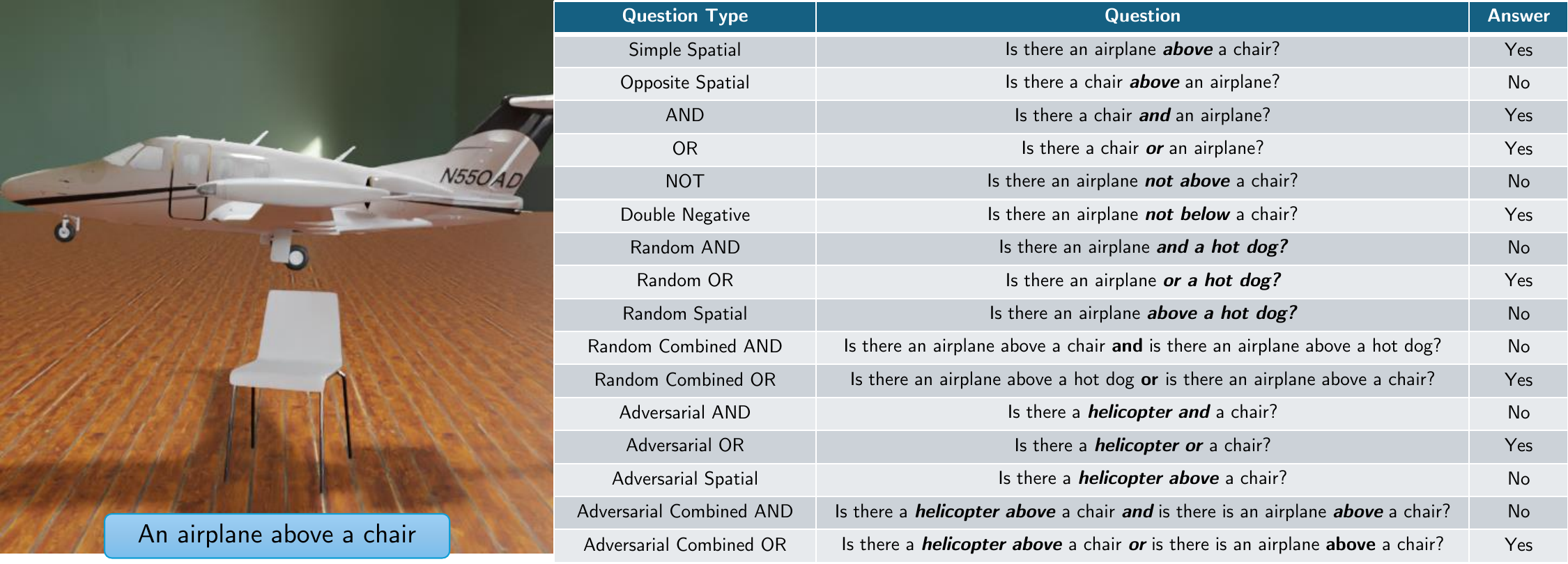}
    \caption{\textbf{The RevQA Benchmark.} Using the REVISION pipeline, we generate spatially accurate images and formulate 16 question types from a given caption. We leverage these generated questions and image, benchmarking Multimodal Large Language Models in their abilities to reason over spatial relationships.}%
    \label{fig:mllm_eval}
\end{figure}

\paragraph{\bf Question Generation.}
 The benchmark contains 16 types of yes-no questions for a REVISION-generated image, consisting of negations, conjunctions, and disjunctions, building on prior work on logic-based visual question answering \cite{gokhale2020vqa}. Each question type evaluates a combination of spatial and logical reasoning abilities in multimodal large language models (MLLMs) (Figure \ref{fig:mllm_eval}). 
 
 Among the 16 types, we incorporate \textit{Random} and \textit{Adversarial} types of questions to further evaluate the robustness and reliability of MLLMs using simple templated transformations. In \textit{Random} types of questions, we replace an object (visible in the image) in the question with another randomly picked object from REVISION's Asset Library. For the \textit{Adversarial} set of questions, we replace one of the objects with another that is semantically and visually close. In addition to benchmarking their robustness, these questions allow simultaneous evaluation of the fine-grained spatial perception and reasoning abilities of these models. To alleviate any order bias in instances which contain multiple questions (see \textit{Combined} in Figure \ref{fig:mllm_eval}), we randomly switch the order between them. 

\begin{table}[t]
\caption{Performances of 5 MLLMs across the 16 types of questions in RevQA. Most models perform worse than random (50\%) when reasoning over Opposite Spatial relationships and Double Negative questions. All models have a significant \textbf{\textit{drop}} in performance with Random/Adversarial questions, in comparison to their simpler versions.}
  \centering
    \resizebox{\linewidth}{!}{
    \begin{tabular}{lccccc}
    \toprule
     {\textbf{Question Type}} & {LLaVa 1.5} & {Fuyu-8B} & {InstructBLIP } & {LLaMA-Adapter 2.1 } & {Qwen-VL Chat} \\
    \midrule
          Simple Spatial & \textbf{0.942} & 0.702 & 0.834 & 0.579 & \underline{0.940} \\
          
          Opposite Spatial & 0.394 & 0.287 & 0.184 & \textbf{0.419} & \underline{0.402}  \\
          
          AND & \underline{0.935} & 0.887 & \textbf{0.957} & 0.858 & 0.889  \\
          
          OR & \textbf{0.995} & 0.396 & 0.598 & 0.722 & \underline{0.949} \\
          
          NOT & \textbf{0.926} & \underline{0.619} & 0.356 & 0.504 & 0.583 \\
          
          Double Negative & 0.267 & 0.347 & \textbf{0.665} & \underline{0.490} & 0.212 \\
          
          Random AND & \underline{0.934} & 0.308 & 0.675 & 0.616 & \textbf{0.978} \\
          
          Random OR & \textbf{0.925} & 0.178 & 0.194 & 0.194 & \underline{0.324} \\ 
          
          Random Spatial & \textbf{0.925} & 0.370 & 0.686 & 0.790 & \underline{0.919} \\

          Random Combined AND & 0.116 & 0.502 & \underline{0.627} & \textbf{0.800} & 0.567 \\

           Random Combined OR & \textbf{ 0.968} & \underline{0.536} & 0.414 & 0.003 & 0.506 \\

           Adversarial AND & \underline{0.661} & 0.184 & 0.542 & 0.641 & \textbf{0.789} \\

          Adversarial OR & \textbf{0.921} & 0.188 & 0.443 & 0.156 & \underline{0.685} \\

          Adversarial Spatial & 0.559 & 0.335 & \underline{0.777} & \textbf{0.893} & 0.615 \\

          Adversarial Combined AND & 0.132 & 0.539 &  \underline{0.695} & \textbf{0.805} & \underline{0.695} \\

          Adversarial Combined OR & \textbf{0.953} & \underline{0.456} & 0.386 & 0.003 & 0.254 \\
          
          \midrule

          \textbf{Average} & \textbf{0.720} & 0.446 & 0.598 & 0.578 & \underline{0.642} \\    \bottomrule
    \end{tabular}%
    }
    \label{tab:mllm_eval_table}%
\end{table}%

\paragraph{\bf Evaluation Setup and Results}
We sample \textit{50k} image-question pairs and benchmark 5 open-source state-of-the-art MLLMs - LLaVA 1.5 \cite{llava}, Fuyu-8B\cite{fuyu-8b}, InstructBLIP\cite{dai2023instructblip}, LLaMA-Adapter 2.1\cite{zhang2023llamaadapter} and Qwen-VL-Chat\cite{bai2023qwenvl}. We instruct all models to generate binary responses and set the temperature $=0$, to remove stochasticity in the generated responses.

We present our evaluation results in Table \ref{tab:mllm_eval_table} and find that all models have a large gap in performance in reasoning over spatial relationships. While most models reason well over simple spatial relationships, they have a large performance drop when presented with the opposite spatial relationships. For example, LLaVA-1.5, the best performing model, has a 58.17\% decrease in performance when probed with \textit{simple} vs \textit{opposite} spatial questions. This can be attributed to : \textbf{\textit{(a)}} insufficient training data for rare object relationships, such as less instances of an ``elephant \textit{above} a person'' than vice versa; \textbf{\textit{b)}} the inability of vision encoders like CLIP to capture subtle semantic differences. MLLMs also struggle with negation, possibly because image captions do not capture enough negations; e.g. COCO Captions only contain\textbf{\textit{ 0.97\%}} occurrences of \textit{'not'}. All models significantly suffer when presented with questions that consist of \textit{double negatives}, which evaluate the models' ability to reason of negations and spatial relationships in tandem. Furthermore, all models suffer under adversarial settings in comparison to their simpler counterparts; comparing LLaVA's performance for \textit{AND} and \textit{Adversarial Combined AND} questions, we find a 85.88\% (0.935\textrightarrow0.132) drop in performance. We also observe a larger decline in performance for \textit{Adversarial} questions than for the \textit{Random} set of questions hinting that while models independently perform well at object recognition and simple spatial relationships, combining them adversarially significantly reduces performance.
\section{Conclusion}

In this work, we introduce REVISION, a framework designed for training-free enhancement of spatial relationships in Text-to-Image models and RevQA, a benchmark to evaluate the spatial reasoning abilities of multimodal large language models. 
Our results demonstrate the effectiveness of leveraging 3D rendering pipelines as a cost-efficient approach for developing generative models with robust reasoning capabilities. 
REVISION is modular and can easily be extended to incorporate additional features, assets, and relationships. We hope our method inspires future research at the intersection of computer graphics and generative AI, enabling safe deployment of these systems in the real world.

\section*{Acknowledgements}

The authors acknowledge resources provided by Research Computing at Arizona State University. The authors also acknowledge technical access and support from ASU Enterprise Technology.  
This work was supported by NSF Robust Intelligence program grants \#1750082 and \#2132724. TG was supported by Microsoft's Accelerating Foundation Model Research (AFMR) program and UMBC's Strategic Award for Research Transitions (START).
The views and opinions of the authors expressed herein do not necessarily state or reflect those of the funding agencies and employers. 

\bibliographystyle{splncs04}
\bibliography{main}

\title{REVISION: Rendering Tools Enable Spatial Fidelity in Vision-Language Models: Supplementary Materials} 

\titlerunning{REVISION}

\author{
Agneet Chatterjee \thanks{Equal contribution. Correspondence to \href{mailto:agneet@asu.edu}{agneet@asu.edu}}\inst{1}\orcidlink{0000-0002-0961-9569} \and
Yiran Luo \textsuperscript{$\star$} \inst{1}\orcidlink{0000-0001-6533-8617} \and \\
Tejas Gokhale \inst{2}\orcidlink{0000-0002-5593-2804} \and 
Yezhou Yang \inst{1}\orcidlink{0000-0003-0126-8976} \and Chitta Baral\inst{1}\orcidlink{0000-0002-7549-723X}
}

\authorrunning{A. Chatterjee et al.}

\institute{Arizona State University \and University of Maryland, Baltimore County}

\maketitle

\setcounter{topnumber}{3}
In this supplementary material, we present additional results on ControlNet and GPT-4 Guided Coordinate Generation results. We also present illustrative samples, covering successful and failure image generation as well as results on the human evaluation experiments. Lastly, we present asset samples from the REVISION pipeline along with outputs from the Position Diversifier.

\section{Additional Quantitative Results}
The ControlNet-based results are presented in Table \ref{tab:controlnet_results}. We achieve the best trade-off between IS and VISOR for ControlNet and compared to Stable Diffusion, we achieve a higher VISOR$_{4}$ score, indicating correctness over multiple trials. Thus, we quantify that REVISION generated images have enough low-level information to faithfully represent spatial orientations.

\begin{table}[t]
\caption{\textbf{ControlNet + REVISION results on VISOR}. As indicated by the high VISOR$_4$ score, we consistently generate images which are spatially correct.}
    \centering
    \small
    \begin{tabular}{@{}ccc rr r cccc@{}}
        \toprule 
          \multirow{2}{*}{\textbf{Background}} & \multirow{2}{*}{\textbf{IS}} & \multirow{2}{*}{\textbf{OA (\%)}} & \multicolumn{5}{c}{\textbf{VISOR (\%)}} \\
         \cmidrule{4-9}
         & & & \textbf{uncond} & \textbf{{cond}} & \textbf{1} & \textbf{2} & \textbf{3} & \textbf{4}\\
        \midrule
        White          &\textbf{18.82}&  \underline{56.88} & 55.48  & \underline{97.54}  & \underline{78.82} & 62.93 & 48.58 & 31.59 \\
        Indoor     &14.75& \textbf{59.64} & \textbf{58.08} & 97.39 & \textbf{81.38}&\textbf{66.35}&\textbf{51.27}&\underline{33.33} \\
        Outdoor       &\underline{16.20}& 56.54&\underline{56.22}&\textbf{99.45}&75.97&\underline{62.99}&\underline{50.57}&\textbf{35.37} \\
        \bottomrule
    \end{tabular}
    
    \label{tab:controlnet_results}
\end{table}
\begin{table}[]
 \caption{\textbf{VISOR Results on using GPT-4  as the Coordinate Generator}. The drop in performance is attributed to the proclivity of GPT-4 to place both the objects too close to each other in the coordinate space.}
    \centering
    \small
    \begin{tabular}{@{}cll rr r cccc@{}}
        \toprule 
        \multirow{2}{*}{\textbf{Method}}  & \multirow{2}{*}{\textbf{OA}} & \multicolumn{2}{c}{\textbf{VISOR}} \\
         \cmidrule{3-4}
         & & \textbf{uncond} & \textbf{{cond}} \\
        \midrule
       SD 1.4 + REVISION &\underline{43.88}&  \underline{41.18} & 93.85 \\
       SD 1.5 + REVISION & \textbf{44.35}&\textbf{41.64}&\underline{93.89} \\
       SD 2.1 + REVISION & 39.03&36.86&\textbf{94.43} \\
        \bottomrule
    \end{tabular}
   
    \label{tab:gpt_results}
\end{table}

\section{GPT-4 Guided Co-ordinate Generation}

W also experiment with generating flexible coordinates for the objects using GPT-4. We first feed GPT-4 a designed in-context prompt that includes specific example coordinates for each possible spatial relation. We then feed in an input prompt of two objects and one spatial relation in order to obtain the two sets of coordinates for placing the mentioned objects. Table \ref{tab:gpt_results} shows results of performing conditioning using GPT-4 as the alternative Coordinate Generator. Compared to our baseline results in, we notice an average of 10-point drop in performance in both Object Accuracy and VISOR$_{uncond}$ scores. These patterns develop as a result of GPT-4's propensity to generate co-ordinates which places the two objects in close proximity, leading to images where the objects are indistinguishable. Hence, T2I models tend to ignore either object, which correspondingly lead to lower VISOR scores.

\section{Object-Wise Spatial Accuracy Analysis}
\begin{figure}[!htp]
    \centering
    \includegraphics[width=\linewidth]{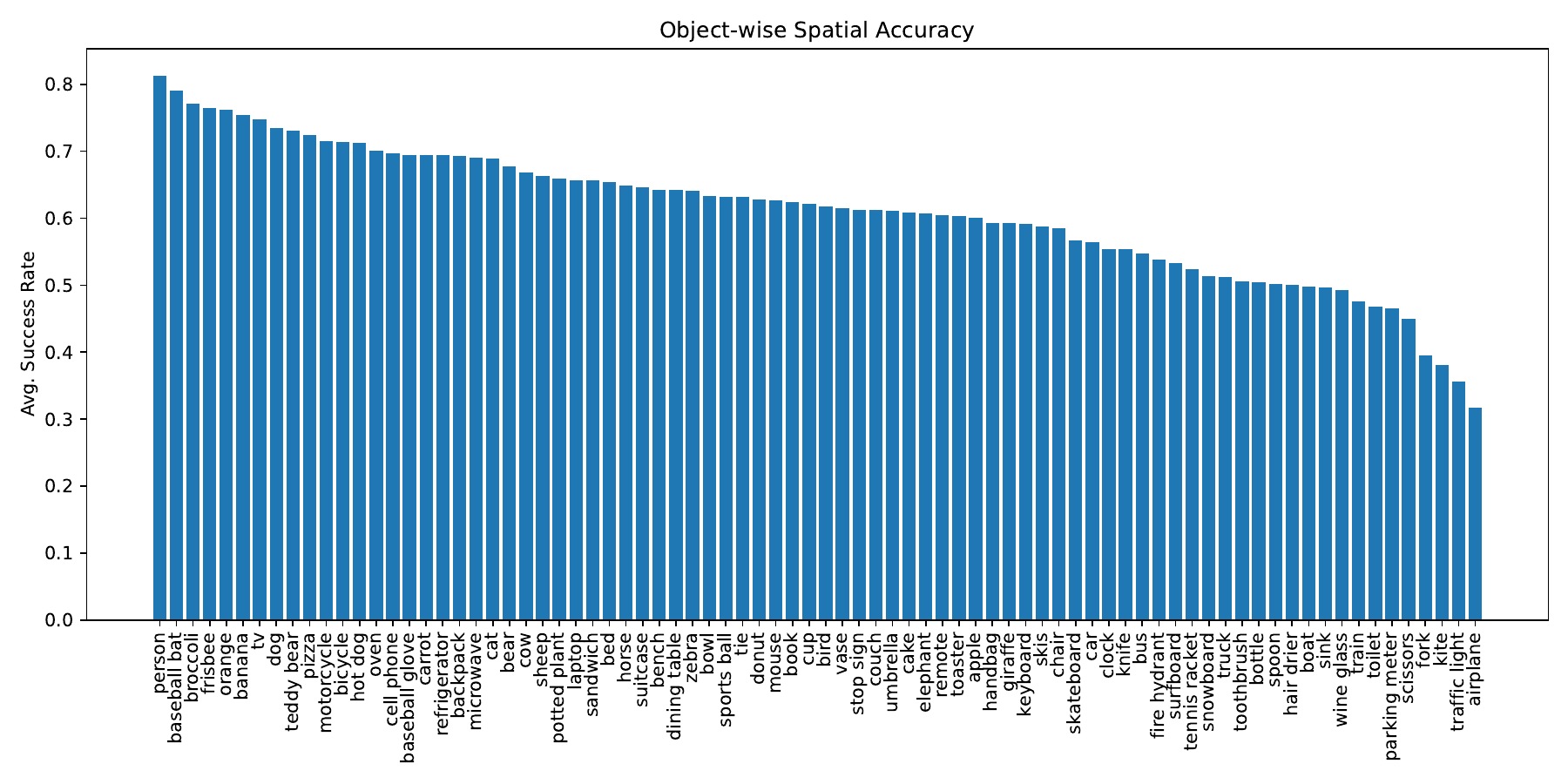}
    \caption{Average Success Rate of each MS-COCO object in REVISION, being spatially correct according to the input prompt in the generated image. We report results using the white background with SD v1.5.}
    \label{fig:obj-wise-success}
\end{figure}

In Figure \ref{fig:obj-wise-success}, we show the success rate of correctly generating objects from MS-COCO using REVISION-based guidance. On average, there is a 61\% likelihood that an MS-COCO object is accurately positioned in the output image.

\section{Illustrative Results for Human Evaluation Experiments}
\subsection{Prompts of Multiple Objects and Relationships} 
\begin{figure}
    \centering
    \includegraphics[width=\linewidth]{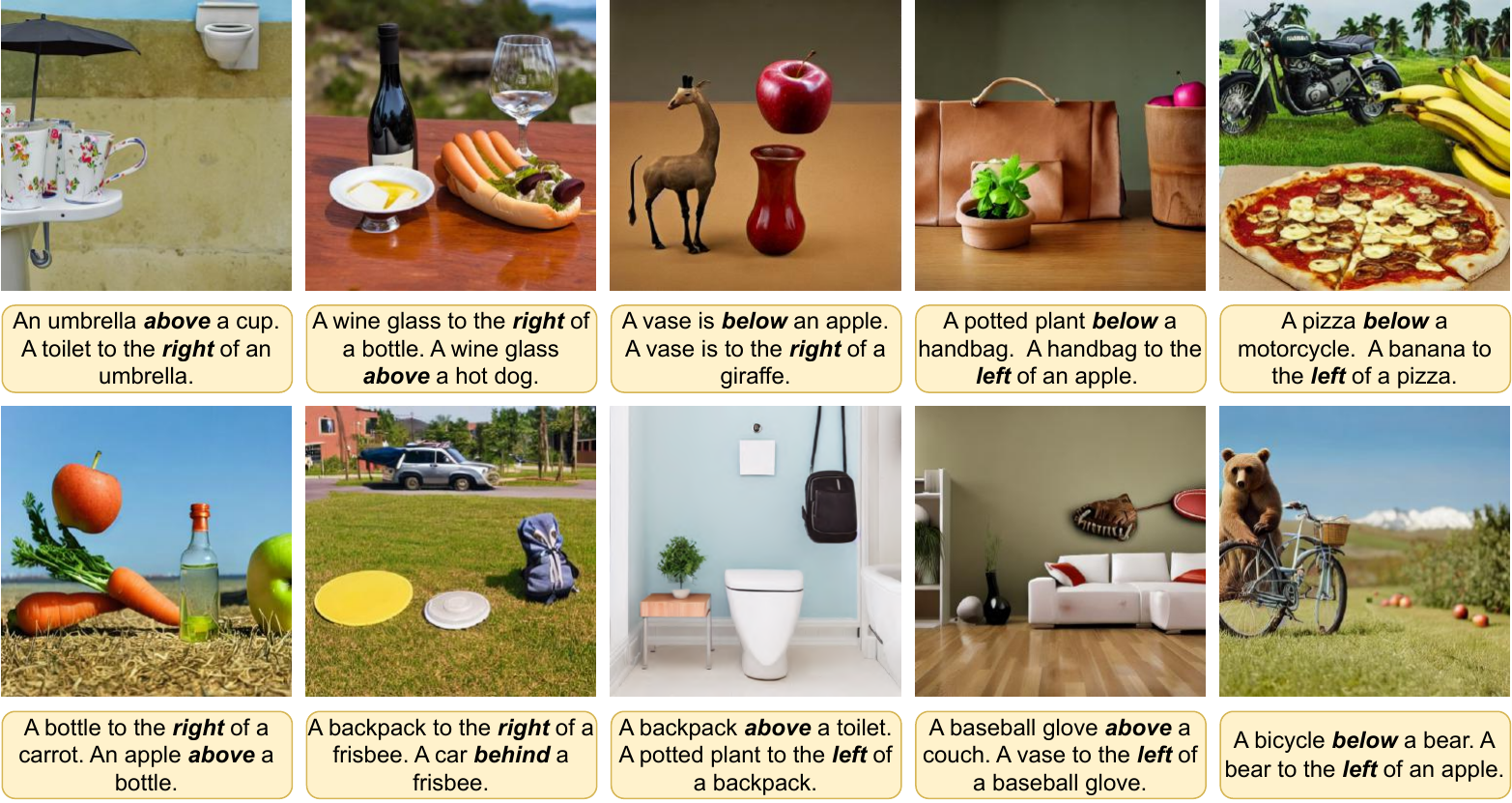}
    \caption{Illustrative example of leveraging REVISION to generate spatially correct images with 3 objects and 2 relationships.}
    \label{fig:eccv_3_obj}
\end{figure}

We present illustrative results in Figure \ref{fig:eccv_3_obj}, where we show that REVISION extends accurate image generation when a prompt includes multiple objects and spatial relationships.

\begin{figure}
    \centering
    \includegraphics[width=\linewidth]{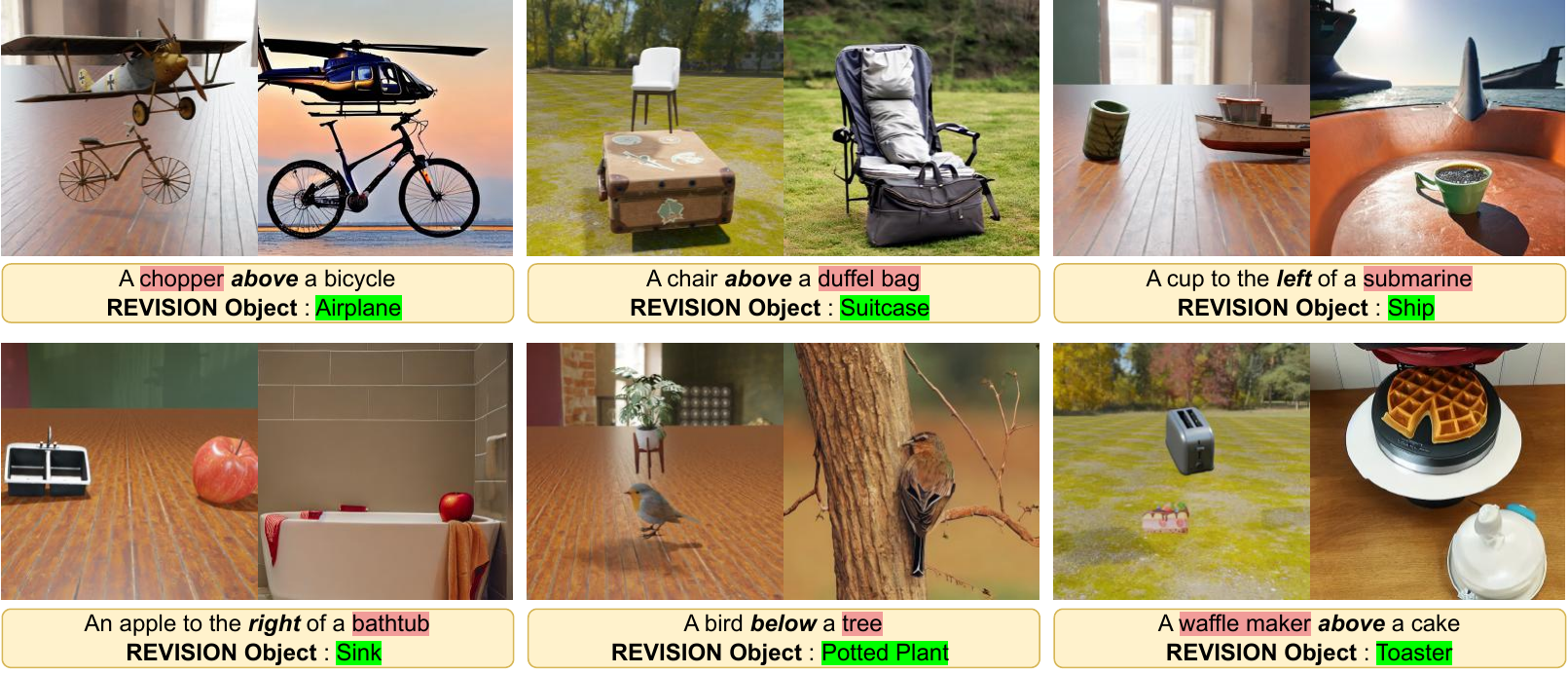}
    \caption{Illustrative example of leveraging REVISION to generate images with objects not in our asset library. For each pair of image, left is the reference image from REVISION, and right is the generated image. Objects in \sethlcolor{green}\hl{green} are from our asset library, while objects in \sethlcolor{pink}\hl{red} are OOD objects.\sethlcolor{green}}
    \label{fig:eccv_ood}
\end{figure}

\begin{table}[!htbp]
\caption{Substitute OOD object nouns for the original 80 MS-COCO objects used in REVISION.}
\centering
\small
\begin{tabular}{lrclr}\toprule
MS-COCO Object&OOD Object & &MS-COCO Object&OOD Object \\\midrule
airplane &helicopter & &kite &flag \\
apple &pear & &knife &sword \\
backpack &purse & &laptop &tablet \\
banana &mango & &microwave &toaster oven \\
baseball bat &walking stick & &motorcycle &tractor \\
baseball glove &boxing glove & &mouse &webcam \\
bear &monkey & &orange &papaya \\
bed &table & &oven &dishwasher \\
bench &sofa & &parking meter &phone booth \\
bicycle &scooter & &person &mannequin \\
bird &butterfly & &pizza &burger \\
boat &submarine & &potted plant &tree \\
book &magazine & &refrigerator &cabinet \\
bottle &lunchbox & &remote &game controller \\
bowl &plate & &sandwich &salad \\
broccoli &cauliflower & &scissors &pliers \\
bus &tram & &sheep &goat \\
cake &pie & &sink &bathtub \\
car &ambulance & &skateboard &roller skates \\
carrot &sweet potato & &skis &hockey stick \\
cat &rabbit & &snowboard &sled \\
cell phone &landline phone & &spoon &straw \\
chair &barstool & &sports ball &bowling ball \\
clock &wall calendar & &stop sign &parking sign \\
couch &cushion & &suitcase &duffel bag \\
cow &panda & &surfboard &kayak \\
cup &tumbler & &teddy bear &doll \\
dining table &dressing table & &tennis racket &badminton racket \\
dog &fox & &tie &bowtie \\
donut &pudding & &toaster &waffle maker \\
elephant &lion & &toilet &shower \\
fire hydrant &mailbox & &toothbrush &comb \\
fork &chopsticks & &traffic light &streetlight \\
frisbee &basketball & &train &roller coaster \\
giraffe &camel & &truck &crane \\
hair drier &hairbrush & &tv &computer monitor \\
handbag &cardboard box & &umbrella &tent \\
horse &donkey & &vase &pitcher \\
hot dog &burrito & &wine glass &glass jar \\
keyboard &piano & &zebra &llama \\
\bottomrule
\end{tabular}

\label{tab: appendix-ood-obj}
\end{table}

\subsection{Out-of-Distribution Objects (OOD)}
For experiments involving OOD objects, we swap one OOD object in the input prompt with its corresponding MS-COCO substitute object. We present the corresponding substitutes in  Table \ref{tab: appendix-ood-obj} and show illustrative examples in Figure \ref{fig:eccv_ood}.

\section{Additional Illustrations}

\begin{figure}[]
    \centering
    \includegraphics[width=\linewidth]{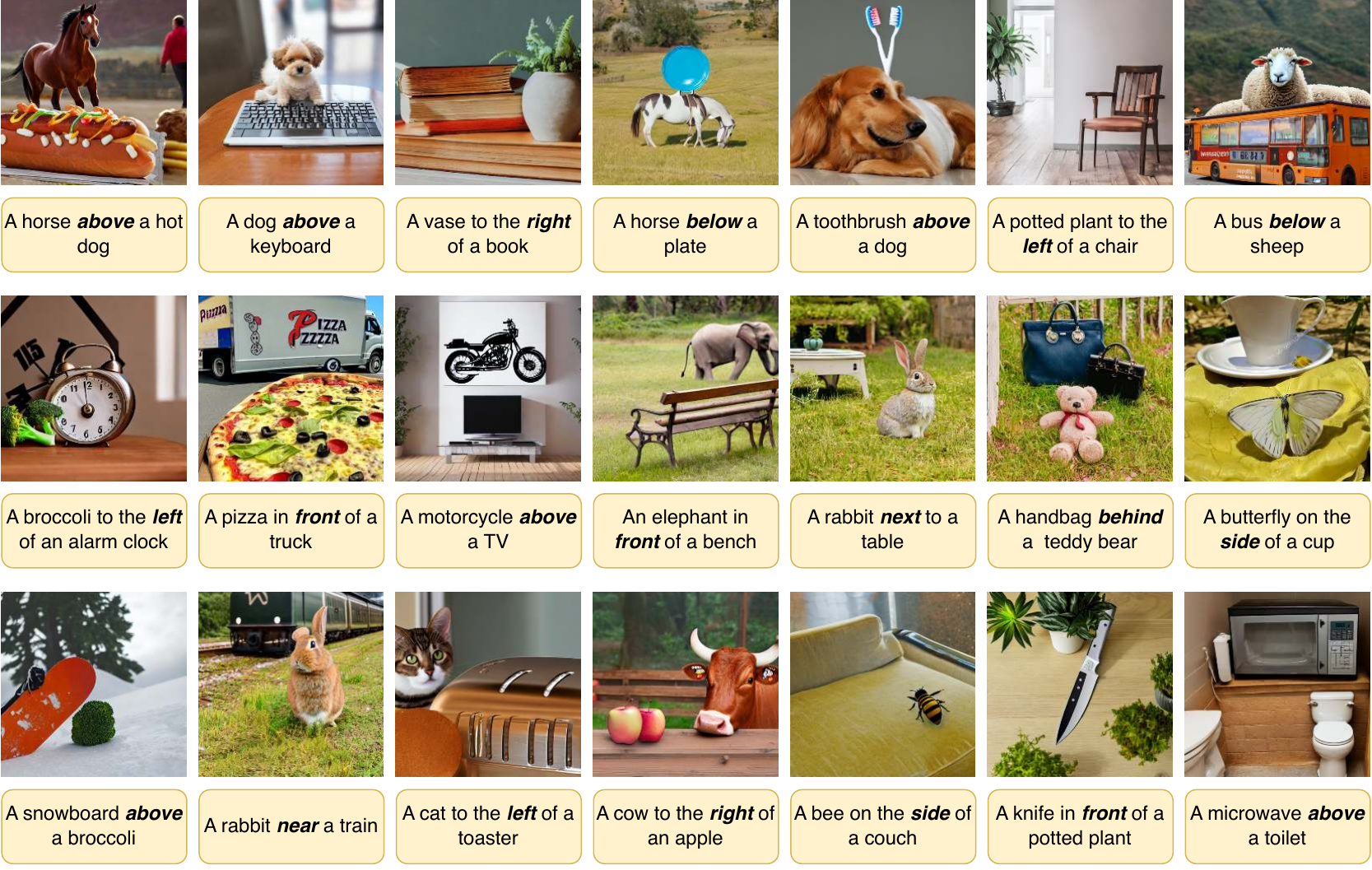}
    \caption{Correctly generated images from T2I models by leveraging images from REVISION as additional guidance.}
    \label{fig:success}
\end{figure}

\begin{figure}[t]
    \centering
    \includegraphics[width=\linewidth]{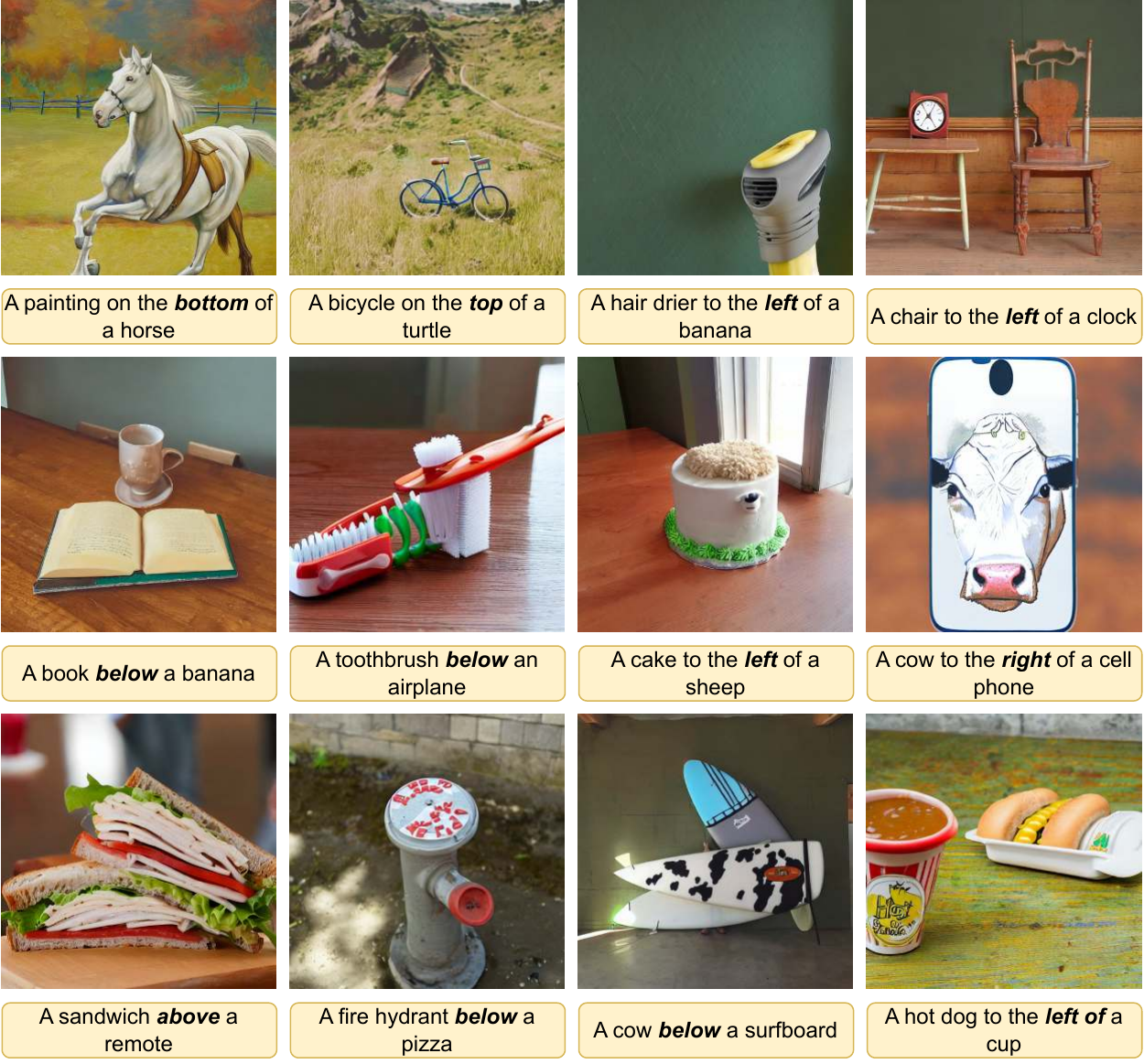}
    \caption{Images generated from T2I models by leveraging REVISION, which either do not contain correct objects or are spatially incorrect.}
    \label{fig:failure}
\end{figure}

Next, we demonstrate additional illustrations of images generated using images from REVISION as additional guidance; Figure \ref{fig:success} shows successfully generated images with REVISION while Figure \ref{fig:failure} presents  failure scenarios. 

\section{REVISION Assets and Illustrations}

\begin{figure}[t]
    \centering
    \includegraphics[width=\linewidth]{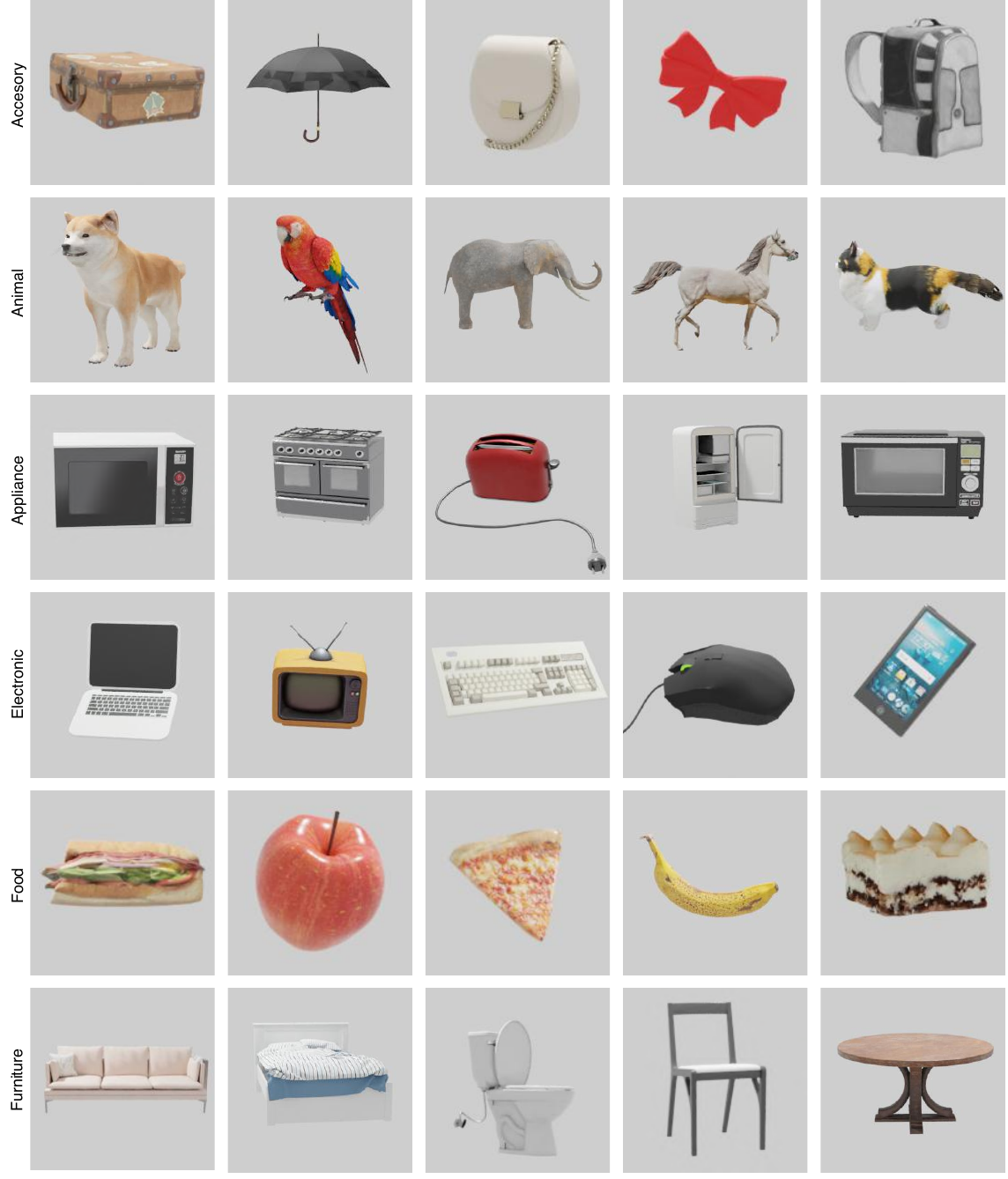}
    \caption{Example 3D models of MSCOCO objects featured in REVISION's Asset Library. }
    \label{fig:appendix-assets-0}
\end{figure}

\begin{figure}[]
    \centering
    \includegraphics[width=\linewidth]{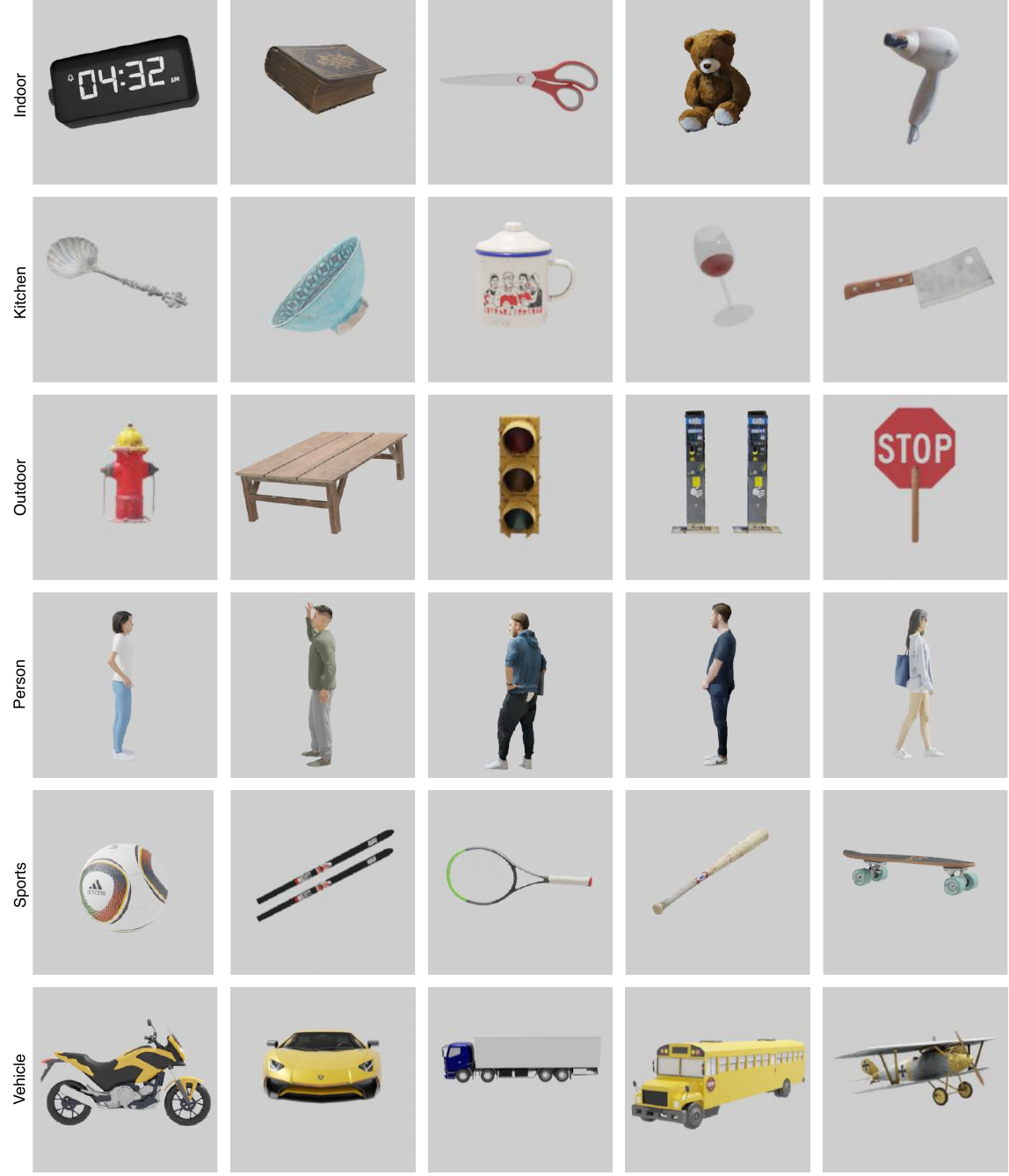}
    \caption{Example 3D models of MSCOCO objects featured in REVISION's Asset Library, continued. }
    \label{fig:appendix-assets-1}
\end{figure}

\begin{figure}
    \centering
    \includegraphics[width=\linewidth]{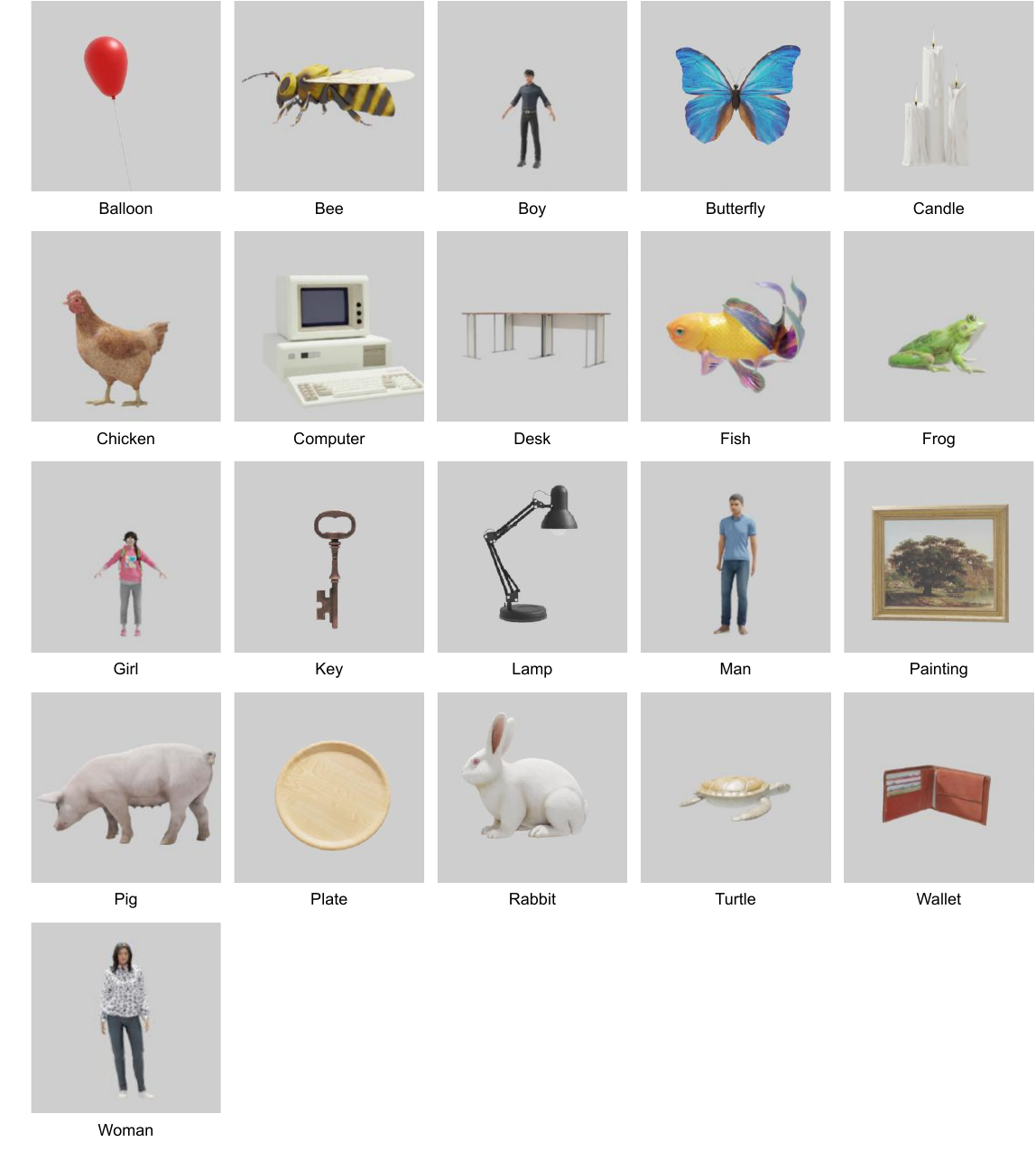}
    \caption{Example 3D models of Non-MSCOCO objects featured in REVISION's Asset Library. }
    \label{fig:appendix-assets-non-ms-coco}
\end{figure}

\begin{figure}[]
    \centering
    \includegraphics[width=\linewidth]{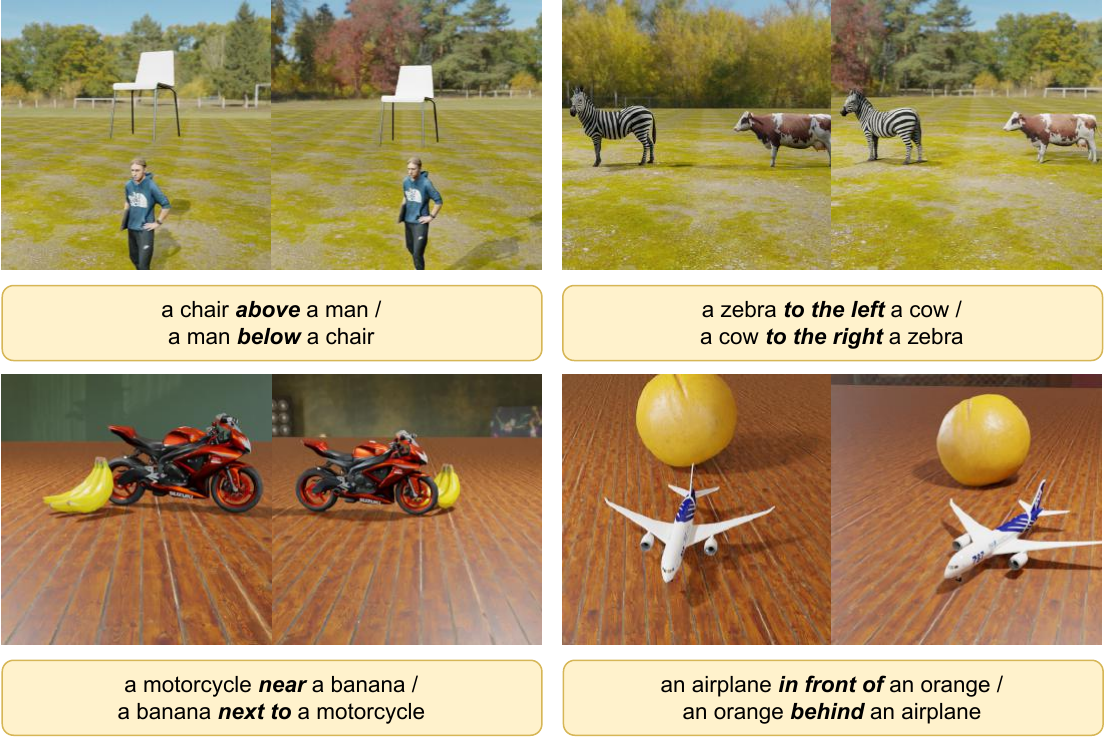}
    \caption{Diversified scenes achieved by the Position Diversifier of REVISION, in all categories of spatial relationships. }
    \label{fig:appendix-assets-ood}
\end{figure}

\begin{figure}[]
    \centering
    \includegraphics[width=\linewidth]{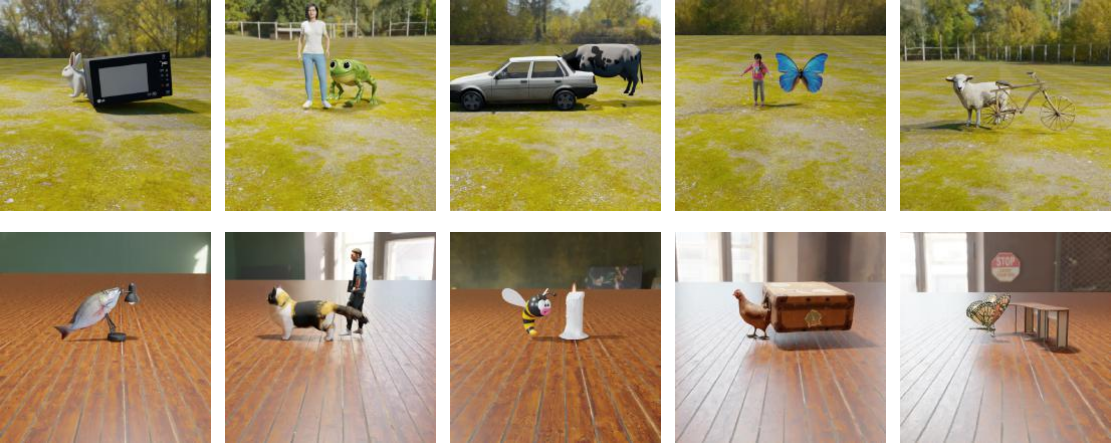}
    \caption{Example REVISION outputs in \textit{near} relationship, featuring two object assets within close proximity or touching each other.}
    \label{fig:appendix-revision-near}
\end{figure}

\begin{figure}[]
    \centering
    \includegraphics[width=\linewidth]{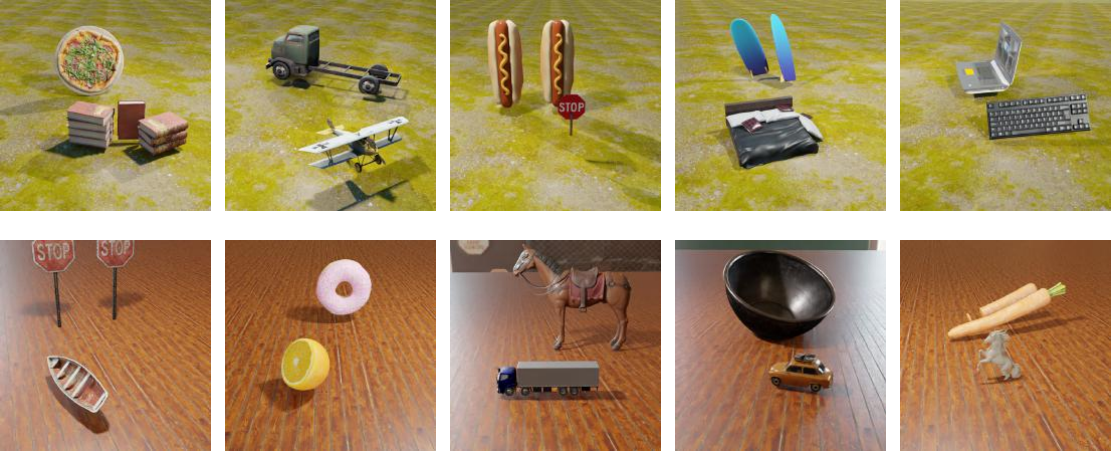}
    \caption{Example REVISION outputs in \textit{depth} relationship, featuring two object assets in front of/behind one another. The angle of the camera is also relatively elevated to strengthen the depth perspective.}
    \label{fig:appendix-revision-depth}
\end{figure}

Figure \ref{fig:appendix-assets-0} and \ref{fig:appendix-assets-1} illustrate instances of MS-COCO 3D assets present in REVISION, organized by subcategory. Figure \ref{fig:appendix-assets-non-ms-coco} presents the non MS-COCO objects and their corresponding class labels.
We present results from our Position Diversifier module in Figure \ref{fig:appendix-assets-ood}. For a given set of assets and spatial relationship, we generate multiple distinct instances.
In Figure \ref{fig:appendix-revision-near} and \ref{fig:appendix-revision-depth}, we present images from REVISION which depict the \textit{near} and \textit{depth} spatial relationships, respectively.



\end{document}